\documentclass[runningheads]{llncs}

\usepackage{eccv}

\usepackage{eccvabbrv}

\usepackage{graphicx}
\usepackage{booktabs}

\usepackage{hyperref}

\newcommand{\bVAE}{$\beta$-VAE}

\newcommand{\scoreest}{\epsilon}
\newcommand{\pcfg}{p_\text{CFG}}

\usepackage{enumitem}
\usepackage{xspace} 
\usepackage{tabularx} 
\usepackage[boldmath]{numprint} 
\npdecimalsign{.}
\nprounddigits{2}

\usepackage{listings}

\newcommand{\name}{GIVT\xspace}
\newcommand{\ours}{\emph{(Ours)}}

\newcommand{\fakeparagraph}[1]{\vspace{1ex}\noindent\textbf{#1\quad}}

\newcommand\blfootnote[1]{%
  \begingroup
  \renewcommand\thefootnote{}\footnote{#1}%
  \addtocounter{footnote}{-1}%
  \endgroup
}

\usepackage{wrapfig}

\begin{document}

\title{GIVT: Generative Infinite-Vocabulary Transformers} 

\titlerunning{GIVT: Generative Infinite-Vocabulary Transformers}

\author{Michael Tschannen \and
Cian Eastwood\inst{*} \and
Fabian Mentzer\inst{\circ}}

\authorrunning{M. Tschannen et al.}

\institute{Google DeepMind \qquad
\email{\{tschannen, mentzer\}@google.com}}

\maketitle

\begin{abstract}
We introduce Generative Infinite-Vocabulary Transformers\\(GIVT) which generate vector sequences with real-valued entries, instead of discrete tokens from a finite vocabulary. To this end, we propose two surprisingly simple modifications to decoder-only transformers: 1) at the input, we replace the finite-vocabulary lookup table with a linear projection of the input vectors; and 2) at the output, we replace the logits prediction (usually mapped to a categorical distribution) with the parameters of a multivariate Gaussian mixture model. Inspired by the image-generation paradigm of VQ-GAN and MaskGIT, where transformers are used to model the discrete latent sequences of a VQ-VAE, we use GIVT to model the unquantized real-valued latent sequences of a $\beta$-VAE.
In class-conditional image generation GIVT outperforms VQ-GAN (and improved variants thereof) as well as MaskGIT, and achieves performance competitive with recent latent diffusion models.
Finally, we obtain strong results outside of image generation when applying GIVT to panoptic segmentation and depth estimation with a VAE variant of the UViM framework.
\blfootnote{$^*$Work done as Student Researcher at GDM. $^\circ$Significant technical contributions. 
Code and model checkpoints:
\href{https://github.com/google-research/big_vision}{https://github.com/google-research/big\_vision}.}
\end{abstract}

\section{Introduction} \label{sec:intro}
After becoming the dominant architecture in natural language processing shortly after their introduction, Transformers~\cite{vaswani2017attention} have also recently become very popular in computer vision~\cite{dosovitskiy2020image, strudel2021segmenter, li2022exploring}.
Dosovitskiy \etal \cite{dosovitskiy2020image} showed that by splitting images into sequences of patches, linearly embedding those patches, and then feeding the resulting sequence of features to a transformer encoder leads to powerful image classifiers that outperform CNN-based architectures at large model and data scale. 
This strategy is now standard for many discriminative vision task including classification~\cite{dosovitskiy2020image}, detection~\cite{li2022exploring}, and segmentation~\cite{strudel2021segmenter}.
It is less obvious how to apply generative transformer decoders to image \textit{generation}
since they were designed to consume and predict \textit{discrete tokens from some fixed, finite vocabulary}. Such a structure naturally fits natural language, for which decoder-only models enable powerful sequential generative modeling and efficient training~\cite{vaswani2017attention, radford2018improving}.

\begin{figure*}[ht]
    \centering
    \includegraphics[width=\linewidth]{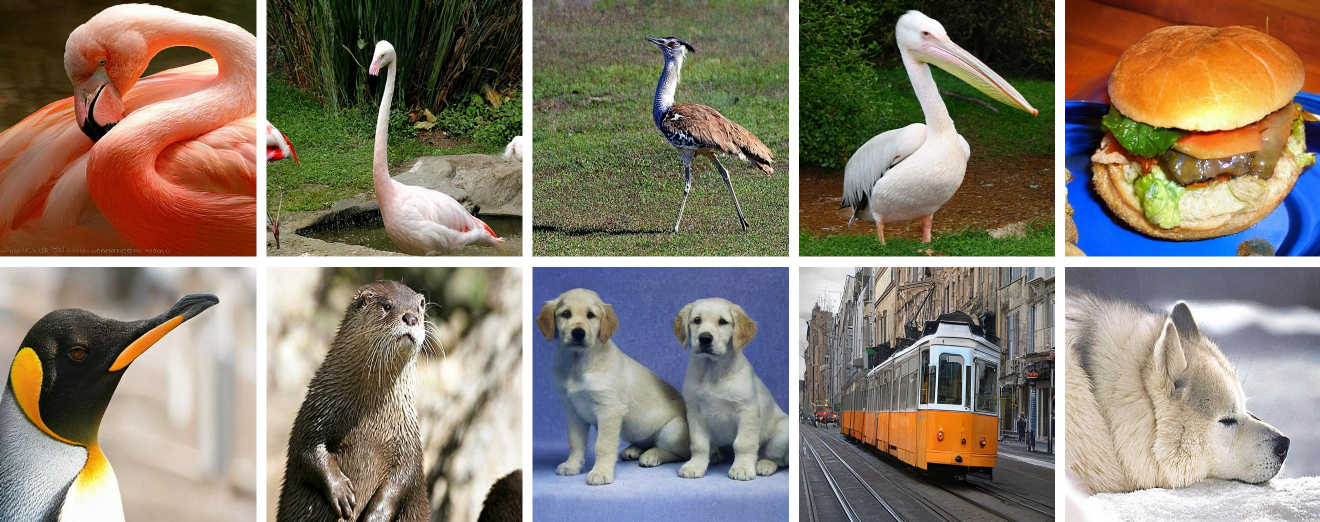}\vspace{-0.2cm}
    \caption{\label{fig:ar_sampling}
    Selected $512\times512$ samples from GIVT-Causal-L for 10 ImageNet classes (130, 130, 138, 144, 933, 145, 360, 207, 829, 248).
    }\vspace{-0.1cm}
\end{figure*}
To harness these capabilities for images, recent works~\cite{razavi2019generating,esser2020taming,chang2022maskgit,chang2023muse,li2023mage,mentzer2023finite} have employed a two-stage approach which first trains a Vector-Quantized Variational Autoencoder (VQ-VAE)~\cite{van2017neural} to map images to a sequence of discrete tokens, and then trains a transformer decoder to model the latent discrete-token distribution.
An advantage of such a VQ-VAE-based image tokenization is that it enables interleaved multimodal generative models, simply by concatenating the vocabularies of the different modalities including text and images~\cite{aghajanyan2022cm3, kim2023magvlt, aghajanyan2023scaling}. However, this approach also has several issues. First, the non-continuous nature of VQ requires differentiable approximations to enable stochastic gradient-based optimization~\cite{van2017neural}. 
Second, a VQ-VAE with a small vocabulary can make the latent modeling easy but also makes the latent code less informative, which prevents control of the low-level details in image generation, and impacts quality when using the tokens for dense prediction~\cite{kolesnikov2022uvim, lu2022unified} or low-level discriminative tasks~\cite{aghajanyan2022cm3, kim2023magvlt}. A large vocabulary, on the other hand, can lead to low vocabulary utilization~\cite{mentzer2023finite} so that high-fidelity VQ-VAE setups typically rely on a range of advanced techniques, such as entropy losses \cite{chang2022maskgit} or codebook-splitting \cite{kolesnikov2022uvim}.
Furthermore, large vocabularies lead to correspondingly large embedding matrices and hence memory consumption, which can be an issue particularly in multimodal contexts.

\begin{figure}[t]
\begin{minipage}[t]{0.48\columnwidth}
\centering
\includegraphics[width=1.05\linewidth]{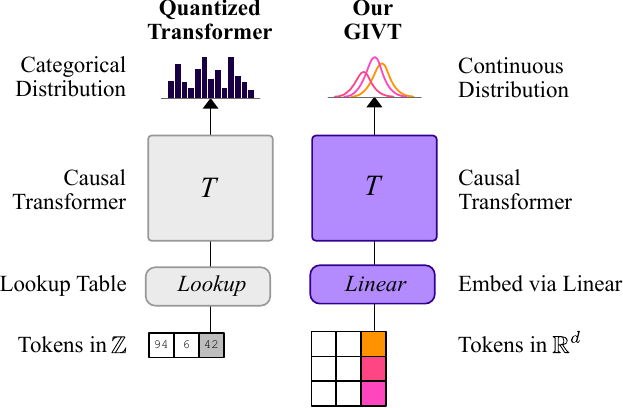}
\caption{\label{fig:teaser}
We compare the standard discrete-token generative transformer (left) to our continuous, infinite-vocabulary variant (\name, right),
using the same decoder-only architecture.
At the input, \name linearly embeds a sequence of \emph{real-valued vectors} instead of discrete tokens via lookup. At the output, \name predicts the parameters of a multivariate, continuous distribution rather than a categorical distribution.}
\end{minipage}\quad\,
\begin{minipage}[t]{0.48\columnwidth}
\centering
\includegraphics[width=\linewidth]{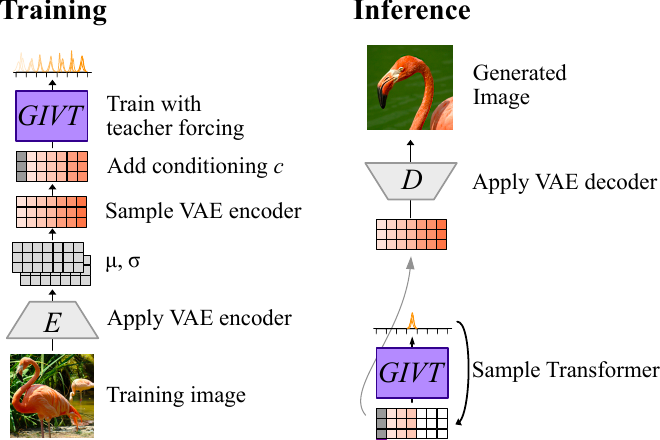}
\caption{\label{fig:arch} GIVT-Causal training and inference. \textit{Left:} During training, we sample a sequence of real-valued latent vectors from the VAE encoder, and train GIVT via teacher forcing. \textit{Right:}
During inference, we sample a sequence of vectors (left-to-right) and feed it to the VAE decoder.
We note that we also explore MaskGIT-like GIVT models not shown here. \emph{No component uses a quantizer.}
}
\end{minipage}
\end{figure}

In this work, we show---to our knowledge for the first time---\textbf{how to completely remove quantization} from generative transformers for visual data.
Indeed, practitioners seem to agree that this would be hardly possible, since transformer decoders are strongly linked to discrete representations in many heads. 
\textbf{Surprisingly, we not only show that simple modifications enable transformer decoders to directly generate sequences of unquantized vectors, but also that this approach leads to better image generation quality and representation learning capabilities than VQ-based approaches.} 
We call such transformers \textit{Generative Infinite-Vocabulary Transformer}~(\name).\footnote{We discuss the relation between continuous latents and infinite vocabulary in Sec.~\ref{sec:additional_discussion}.} Concretely, we make two changes compared to the standard transformer decoder architecture~\cite{vaswani2017attention, radford2018improving}, see \cref{fig:teaser}: 1) at the input, rather than using a sequence of discrete tokens to look up a finite vocabulary of embeddings,
GIVT linearly embeds a sequence of real-valued vectors; and 2) at the output, rather than predicting a categorical distribution over a finite vocabulary, GIVT predicts the parameters of a $d$-variate Gaussian Mixture Model (GMM).
We train GIVT in the same way as standard transformer decoders: with a causal attention mask and teacher forcing~\cite{vaswani2017attention}, and alternatively also explore fast progressive masked-bidirectional-modelling as in MaskGIT~\cite{devlin2018bert,chang2022maskgit,chang2023muse}.

Similar to the two-stage approach with VQ-VAEs and analogous the two-stage approach of latent-diffusion models~\cite{rombach2021high,peebles2022scalable}, we first learn a lower-dimensional latent space with a Gaussian-prior $\beta$-VAE~\cite{kingma2013auto,higgins2016beta}, and then model it with GIVT. We emphasize that training both $\beta$-VAE and GIVT only relies on standard techniques from the deep-learning toolbox, and not the advanced training techniques of the VQ-VAE literature like auxiliary losses~\cite{van2017neural, chang2022maskgit} on the latent representation, codebook reinitialization~\cite{lancucki2020robust}, or dedicated optimization algorithms~\cite{kolesnikov2022uvim, huh2023straightening}.

\noindent Our main contributions can be summarized as follows:
\begin{enumerate}
    \item We show that GIVT outperforms VQGAN~\cite{rombach2021high} (and follow-up variants) and MaskGIT~\cite{chang2022maskgit} in class-conditional image generation, often by a large margin and/or at significantly lower computational cost. GIVT is also competitive with strong latent diffusion baselines, particularly at high resolution.
    \item We derive variants of standard sampling
    techniques for the continuous case, such as temperature sampling, beam search, and classifier-free guidance~(CFG)\cite{ho2022classifier}, and showcase their effectiveness. 
    \item We demonstrate that GIVT matches or outperforms prior sequential image generation models in representation learning at significantly lower computational cost.
    \item GIVT achieves comparable performance with the VQ-based UViM approach \cite{kolesnikov2022uvim} in dense prediction tasks like semantic segmentation and monocular depth estimation.
\end{enumerate}
We emphasize that advances in transformer decoder-based models for visual data generation as GIVT directly benefit form advances in scaling and inference efficiency for large language models. Conversely, and unlike for diffusion models, improvements in models as ours are straight-forward to transfer to multimodal interleaved modeling~\cite{aghajanyan2022cm3, kim2023magvlt, aghajanyan2023scaling} which is becoming increasingly popular.

\section{Related work} \label{sec:relwork}

\fakeparagraph{VQ-VAE for visual data tokenization} Following the success of pixel-space autoregressive modeling~\cite{van2016pixel, salimans2016pixelcnn, parmar2018image, menick2018generating, chen2020generative} for image generation, moving the autorgressive modeling to the latent space of VQ-VAEs~\cite{van2017neural, razavi2019generating} emerged as a more efficient alternative. The use of GANs and perceptual losses for VQ-VAE training as well as modern causal~\cite{esser2020taming, yu2022scaling, villegas2022phenaki} and masked~\cite{chang2022maskgit, chang2023muse, li2023mage} transformers for latent modeling led to substantial quality improvements. Another active area leveraging VQ-VAEs is interleaved multimodal generative modeling of images and text~\cite{aghajanyan2022cm3, kim2023magvlt, aghajanyan2023scaling}. Further, VQ-VAEs are a popular choice to tokenize the label space of dense prediction vision tasks~\cite{kolesnikov2022uvim, lu2022unified}. Finally, some language-inspired techniques for self-supervised learning from images rely on VQ-VAE representations~\cite{bao2021beit,wang2022bevt, li2023mage}.

\fakeparagraph{Discretized mixtures of distributions} replace the dense prediction of the logits of a categorical distribution with a continuous mixture model which is subsequently discretized. This approach was proposed in~\cite{salimans2016pixelcnn} for pixel-space autoregressive modeling, to reduce the number of model parameters and to improve learning efficiency, and is also popular in neural compression \cite{mentzer2020learning, cheng2020learned, mentzer2023m2t}.

\fakeparagraph{Continuous outputs in NLP} A popular approach to handle large vocabularies in machine translation is to predict language tokens via their word embeddings with a continuous distribution, instead of token IDs with a categorical distribution~\cite{kumar2018mises, kumar2021machine, tokarchuk2022target, tokarchuk2023unreasonable,li2019efficient}. Decoding is usually done in greedy fashion with embedding lookup and hence does not produce diverse samples. Further, the models consume and predict word embeddings form a fixed, finite set.

\fakeparagraph{VAEs with learned priors} A rich body of literature studies improving VAEs with learned priors: Inverse autoregressive flows emerged as a popular choice~\cite{kingma2016improved, chen2016variational}. Other approaches use normalizing flows~\cite{vahdat2020nvae} or a mixture of variational posteriors with pseudo-inputs \cite{tomczak2018vae}. For VAEs with discrete (non-VQ) latents, learned priors based on Restricted Boltzmann Machines were studied \cite{vahdat2018dvae, sadeghi2019pixelvae}.

\fakeparagraph{Time-series modeling with Transformers}
A variety of works has recently explored transformers for time-series modeling/forecasting. Those works either use a regression loss~\cite{zhou2021informer, nie2022time, kunz2023deep, garza2023timegpt, das2023decoder}, quantile forecasting~\cite{eisenach2020mqtransformer, lim2021temporal}, or resort to discretizing/binning the data~\cite{rasul2023lag}. Somewhat related, \cite{nachmani2023lms, chen2023lauragpt} regress continuous speech features from discrete tokens. None of these models predict a continuous distribution like \name that allows for autoregressive generation.

\section{Generative infinite-vocabulary transformers}\label{sec:method}

As mentioned in Sec.~\ref{sec:intro},
our method is conceptually similar to recent works that train decoder-only transformer models on the discrete codes of VQ-VAEs~\cite{esser2020taming,yu2021vector,chang2022maskgit,chang2023muse}, with the crucial difference being that we do not quantize (\ie, do not use VQ). We now describe the components of our method.

\subsection{VAE training}\label{sec:method:vae}

We first train a \emph{continuous-latent} \bVAE~\cite{higgins2016beta} with Gaussian encoder and prior as originally proposed by \cite{kingma2013auto}. Given an input image $x$,
the encoder $E$ predicts mean $\mu$, and covariance $\sigma$ of a multivariate normal distribution with diagonal covariance matrix,
and samples a representation $z$ from $\mathcal N(\mu, \sigma)$ using the reparametrization trick~\cite{kingma2013auto}. The VAE decoder then maps the latent sequence back to an image.
Since we use a Gaussian encoder distribution, the KL-term in the evidence lower bound (ELBO)~\cite{kingma2013auto} can be computed in closed form as described in \cite[Sec.~F.1]{kingma2013auto}. As for the reconstruction/likelihood term in the ELBO, we rely on a mixture of MSE, perceptual loss and GAN loss for image generation following \cite{esser2020taming,chang2022maskgit}, or the categorical cross-entropy for dense prediction tasks~\cite{kolesnikov2022uvim}.
Our encoder spatially-downsamples $x$, whereby we obtain $z$ with spatial dimensions $h \times w$ and feature dimension $d$, with $ h{=}\lceil H/16\rceil, w{=}\lceil W/16\rceil$, given a $H{\times}W$ input $x$. To compute the KL-term, the associated $\mu$ and $\sigma$ with shapes $w \times h \times d$ are flattened into $whd$ vectors.

The hyperparameter $\beta$ multiplying the KL-term controls how strongly $z$ is regularized. As we shall see in Sec.~\ref{sec:results}, this regularization of the VAE is important to be able to model the resulting (true) latent distribution $p(z)$ well.
\begin{table*}[t]
\caption{\label{tab:fid_results}
    Results on class-conditional $256{\times}256$ ImageNet, where GIVT-Causal models outperform their quantization-based counterparts at much smaller model size (VQGAN) or substantially shorter sequence length (ViT-VQGAN).
    We report FID as well as precision and recall (where available). We use the standard ADM evaluation suite, where FID is calculated w.r.t.\ the training set.
      \emph{+A}: GIVT variants with adapter,
      \emph{CG}: Classifier guidance acceptance rate or scale,
      \emph{CFG $=w$}: Classifier-free guidance with weight $w$~\cite{ho2022classifier},
      \emph{DB-CFG $=w$}: Our distribution based CFG variant (Sec.~\ref{sec:dbcfg}),
      \emph{Top-k}: Top-k sampling~\cite{fan2018hierarchical} (``mixed'' refers to multiple $k$),
      \emph{$t$}: Temperature sampling by scaling the predicted $\sigma$ of our models with $t$,
      \emph{$t_C$}: Choice temperature for MaskGIT.
      \emph{Steps} number of inference steps.
      Additional comments:
      $^\dagger$Numbers obtained by us from public code,
      $^\star$Inference uses activation caching.
     }
    \centering
    \scriptsize
    \begin{tabular}{lllr%
    n{2}{2}%
    n{2}{2}%
    n{2}{2}%
    }
    \toprule 
    & Model            & Inference & {Steps} &
       \multicolumn{1}{c}{FID$\downarrow$} & 
       \multicolumn{1}{c}{Precision$\uparrow$} & 
       \multicolumn{1}{c}{Recall$\uparrow$} \\
    \midrule
    GANs%
        & BigGAN-deep~\cite{brock2018large}
                       &                  & & 6.95 & {\npboldmath} 0.87 & 0.28 \\
        & StyleGAN-XL~\cite{sauer2022stylegan}
                       &                  & & {\npboldmath}2.30  & 0.78 & {\npboldmath} 0.53  \\
    \midrule
    Diffusion%
        & ADM~\cite{dhariwal2021diffusion}
                       &                  & 250\phantom{$^\star$} & 10.94 & 0.69 & 0.63 \\
    Models    & ADM-G~\cite{dhariwal2021diffusion}
                       & CG $=1.0$          & 250\phantom{$^\star$} & 4.59  & 0.82 & 0.52 \\
        & LDM-4~\cite{rombach2021high} & & 250\phantom{$^\star$} &
        10.56 & 0.71 & 0.62 \\
        & LDM-4-G~\cite{rombach2021high} & CFG $=1.5$ & 250\phantom{$^\star$} &
        3.60 & {\npboldmath} 0.87 & 0.48 \\
        & DiT-XL/2~\cite{peebles2022scalable}
                       &                  & 250\phantom{$^\star$} & 9.62 & 0.67 & {\npboldmath}0.67 \\
        & DiT-XL/2-G~\cite{peebles2022scalable}
                       & CFG $=1.5$         & 250\phantom{$^\star$} & {\npboldmath}2.27 & 0.83 & 0.57 \\
    \midrule
Masked
            & MaskGIT~\cite{chang2022maskgit}
                      & $t_C=4.5$        & 16\phantom{$^\star$}  & 4.916 $^\dagger$ & 0.836$^\dagger$ & {\npboldmath} 0.489$^\dagger$ \\
Modeling
                & \name-MaskGIT  \emph{(Ours)}
                      & $t_C=35$ & 16\phantom{$^\star$}  &
                      4.64 & 0.85 & {\npboldmath} 0.49	 \\
        & \name-MaskGIT  \emph{(Ours)}
                      & $t_C=60$, DB-CFG $=0.1$ & 16\phantom{$^\star$} &
                      {\npboldmath} 4.53 & {\npboldmath} 0.87 & 0.47	\\
\midrule
Sequence
        & VQGAN~\cite{esser2020taming}
                       & Top-k $=$ Mixed      & 256$^\star$ &
                       17.04 \\
Models
        & VQGAN~\cite{esser2020taming}      
                       & Top-k $=600$, CG $=0.05$ & 256$^\star$ &
                       5.20 \\
        & ViT-VQGAN-L~\cite{yu2021vector}
                       & & 1024$^\star$ &
                       4.17 \\
        & ViT-VQGAN-L~\cite{yu2021vector}
                       & CG $=0.5$ & 1024$^\star$ &
                       3.04 \\
        & \name-Causal   \emph{(Ours)}
                      & $t=0.9$         & 256$^\star$ &
                      5.6729 & 0.7490 & 0.5927 \\
        & \name-Causal   \emph{(Ours)}
                      & $t=0.95$, DB-CFG $=0.4$ & 256$^\star$ & 
                      3.35 & {\npboldmath} 0.84 & 0.53 \\
        & \name-Causal-L+A   \emph{(Ours)}
                      & $t=0.9$ & 256$^\star$ & 
                      3.4556 & 0.7701 & {\npboldmath} 0.6131 \\
        & \name-Causal-L+A   \emph{(Ours)}
                      & $t=0.95$, DB-CFG $=0.4$ & 256$^\star$ & 
                      {\npboldmath} 2.5933 & 0.8085 & 0.5695 \\
    \bottomrule
    \end{tabular}
\end{table*}

\subsection{\name training}\label{sec:method:stage2}

We next train a GIVT to predict $p(z)$ or $p(z | c)$ (when a conditioning signal $c$ is  available, \eg, in class-conditional generation).
The representation $z$ is reshaped into a $hw$-length sequence of $d$-dimensional \emph{real-valued} vectors (or ``soft tokens'').
Note how this differs from the standard VQ-VAE-based setup, where the latent transformer decoder models a $hw$-length sequence of \textit{integers} denoting codebook indices.
To accommodate this difference, we make two small changes to the standard transformer decoder-only architecture (see Fig.~\ref{fig:teaser}):
We replace the embedding lookup tables at the input with a single linear layer to project from $d$ to the transformer hidden dimension.
At the output, we do not predict a categorical distribution, and instead let the transformer predict the parameters of a continuous distribution.
Assuming channel-wise independence of the mixture components, we model this continuous distribution with a $k$-mixture GMM. The GIVT model hence predicts $2 k d + k$ parameters per soft token ($kd$ mean and $kd$ variance parameters for the mixture components, and $k$ mixture probabilities).
Experimentally, we found it beneficial to normalize the mixture probabilities with a softmax activation, and the variance parameters with softplus. 

We use the standard cross-entropy loss (which is equivalent to the negative log-likelihood)
on the distribution $\tilde p$ predicted by GIVT, and minimize 
    $\mathcal L_\text{T} = \sum_c \mathbb E_z \left[
        -\log \tilde p(z | c)
    \right]$,
assuming the the classes or conditioning signal $c$ uniformly distributed (see App.~\ref{app:loss_function} for details on the loss).
We train two types of GIVT models, as described next.

\fakeparagraph{GIVT-Causal} Here, GIVT is trained to predict every $d$-dimensional vector in the $hw$ sequence of latents conditioned on all previous vectors. Thereby, the self-attention layers are masked to be temporally causal~\cite{esser2020taming,vaswani2017attention} (which enables sequential generation at inference time and is unrelated to causal inference). 
This training strategy is also called teacher forcing and is analogous to the latent modeling in VQ-GAN~\cite{esser2020taming}. For class-conditional image generation we prepend a \texttt{[CLS]} vector to the input sequence, \ie, a learned vector for each class $c$.

\fakeparagraph{GIVT-MaskGIT} As in MaskGIT~\cite{chang2022maskgit},
we mask a subset of the input sequence randomly during training and then gradually uncover the masked tokens during inference. The only changes compared to~\cite{chang2022maskgit} are related to our real-valued tokens:
since we have infinitely many tokens, there is no obvious choice to define a special mask token (when using VQ, one can just extend the vocabulary to contain special tokens, such as \texttt{[MASK]}).
Instead, given $z$ and a mask $M$ indicating for every location whether it is masked, we first replace the locations in $z$ corresponding to $M$ with zeros (to remove information),
and then embed it with a single dense layer, as above. Additionally, we \emph{concatenate} one of two learned special vectors in the feature dimension, a \texttt{[MASK]} vector for masked locations, and a \texttt{[UNMASK]} vector otherwise (we half the dimension of the embedded inputs and special tokens s.t.\ the final hidden dimension remains unchanged).

\subsection{Towards end-to-end training: Adapters}

An interesting consequence of using an unquantized VAE and modeling the resulting latent sequence with a continuous rather than a categorical distribution is that the VAE and GIVT can be jointly trained or fine-tuned end-to-end (using the reparametrization trick~\cite{kingma2013auto}). However, this setup comes with its own set of challenges (\eg, it encompasses multiple losses which have to be balanced appropriately) and we leave it for future work. Instead, we explore a simple alternative to better match the latent distributions of the VAE and the one predicted by GIVT: We use a small invertible flow model~\cite{dinh2014nice, dinh2016density}, or ``adapter'', to map the VAE latent sequences to a new latent space of identical dimensions. We rely on a ``volume preserving'' additive coupling layer-based model which has a diagonal Jacobian~\cite{dinh2014nice}.
GIVT is then trained jointly with the adapter to predict the sequences in this transformed latent space induced by the adapter (using the same loss). At inference time, samples drawn from GIVT are first processed by the inverted adapter, and then decoded to an image with the VAE decoder. Note that the adapter does not require additional losses thanks to invertibility and adds a negligible compute and model parameter overhead (less than 0.1\%) compared to the GIVT model (see Sec.~\ref{sec:experiments} and App.~\ref{sec:architecture_details} for details).

\subsection{Inference}\label{sec:method:inference}

Given a VAE and GIVT trained as above, during inference 
we sample form GIVT either sequentially (see Fig.~\ref{fig:arch}) or as in MaskGIT~\cite{chang2022maskgit} and decode the sampled sequence into an image.
We now investigate the various inference schemes for discrete transformers, and derive their continuous counterparts.

\fakeparagraph{Temperature Sampling, Nucleus Sampling, Beam Search}\label{sec:method:inference:techniques}
In sequence models for text (see \cite{holtzman2019curious} for an overview and discussion) and VQ-GAN-based approaches, it is common to adapt and tune the sampling algorithm. We start with temperature sampling, which for discrete models adapts the softmax temperature of the categorical distributions predicted at each decoding step.
For GIVT, we instead scale the covariance matrices of the predicted Gaussian distributions and call this strategy ``variance scaling''. As we will see in \cref{sec:experiments}, this simple change can have a significant impact on sample quality.

Nucleus sampling \cite{holtzman2019curious} proposes to collect the largest logits such that its cumulative probability after normalization exceeds a threshold (for example 0.8), and to sample from this reduced-support distribution. In GIVT, when predicting a single mixture, this can be approximated by truncating the predicted distributions per dimension (thereby choosing a higher-density support). This has a similar effect to variance scaling and therefore do not pursue this strategy.

We also consider beam search, which is the same for GIVT as it is for discrete transformer decoders. For every sample, we maintain $B$ beams, and at every step we sample a number of candidates for every beam (we call these ``fans'' here). We then compute the cumulative log probability for all beams and fans up to the current sampling step, and select the $B$ beams with the highest cumulative log probability. Finally, there is no analogous concept for top-k sampling~\cite{fan2018hierarchical} in GIVT, because it predicts continuous distributions.
\begin{figure}[t]
\begin{minipage}[t]{0.45\columnwidth}
    \centering
    \includegraphics[width=\linewidth]{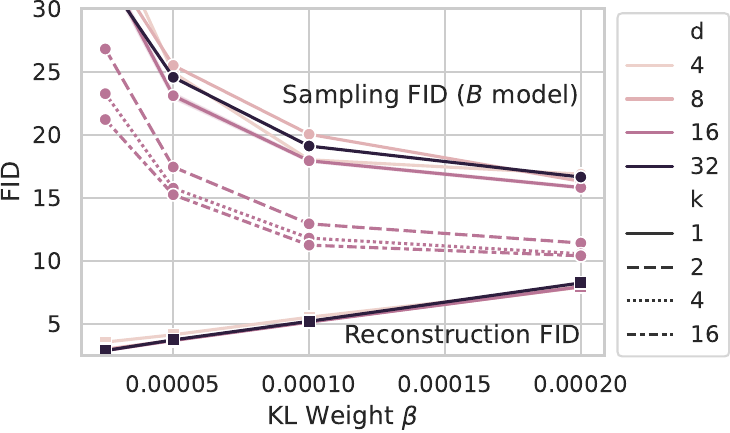}
    \caption{\label{fig:kl_abl}$\beta$-VAE ablation: Interplay of KL weight $\beta$, number of channels $d$, and number of mixtures $k$ when training the VAE. Round markers show the sampling FIDs obtained with a Base-size GIVT-Causal. As $\beta$ and $k$ increase, the sampling FID improves, but the reconstruction FID also increases, limiting the best possible sampling FID. 
    }
\end{minipage}\quad
\begin{minipage}[t]{0.54\columnwidth}
    \centering
    \includegraphics[width=\linewidth]{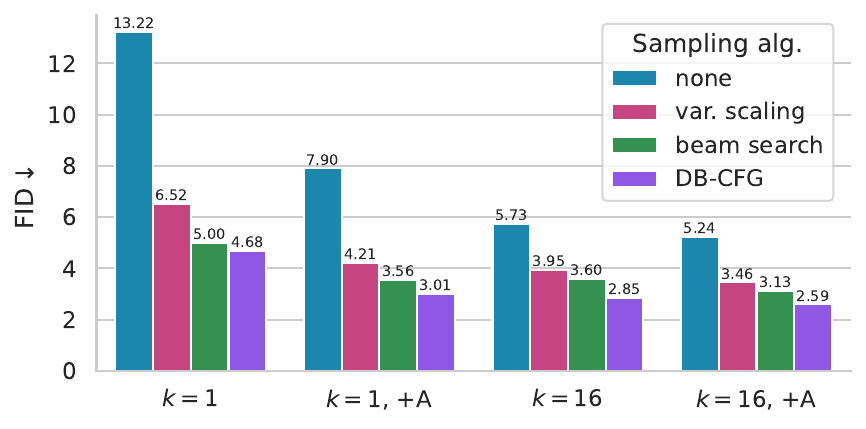}
    \caption{\label{fig:sampling_alg_abl} Effect of different sampling strategies and model variants (GIVT-Causal-L) on sample quality. Increasing the number of mixtures $k$ and adding an adapter (+A) lead to compounding improvements. DB-CFG is the most effective sampling strategy for all model configurations.
    }
\end{minipage}
\end{figure}

\begin{wrapfigure}{r}{5.2cm}
\vspace{-0.5cm}
    \centering
    \includegraphics[width=\linewidth]{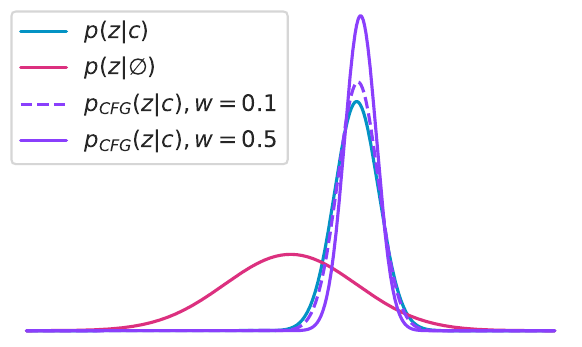}
    \caption{\label{fig:cfg_highlevel}
      Visualization of our \emph{Density-Based Classifier-Free Guidance~(DB-CFG)}.
      We show the conditional and unconditional PDFs predicted by GIVT, and the resulting CFG PDF for two values of $w$.
      Note how the CFG distributions become more peaked. We use rejection sampling to sample from $p_\text{CFG}$.}
\vspace{-0.5cm}
\end{wrapfigure}

\fakeparagraph{Distribution-Based Classifier-Free Guidance}\label{sec:dbcfg}
In the diffusion literature, classifier-free guidance (CFG)~\cite{ho2022classifier} has been employed with great success.
Concretely, conditional diffusion models are trained with an additional null class $\emptyset$ to learn the unconditional data distribution.
Then, during inference, the conditional log density is ``moved away'' from the unconditional one: given a guidance weight $w$, the updated (diffusion) score estimate is is obtained as
\begin{equation}
\label{eq:cfg}
\tilde \scoreest(z, c) = (1 + w) \scoreest(z, c) - w \scoreest(z, \emptyset),
\end{equation}
where $\scoreest$ estimates the gradient of the log density of the data distribution, 
$\scoreest(z,c) \propto \nabla_{z} \log \tilde p(z|c)$ (see~\cite[Sec. 2]{ho2022classifier}).
From this, we now derive a CFG variant for our GIVT, since \emph{we directly predict a density}.
We term this approach ``Density-Based CFG'' (DB-CFG).
Eq.~\ref{eq:cfg} can be written as
\begin{align*}
\tilde \scoreest(z,c) 
  &\propto (1+w) \nabla_{z} \log \tilde p(z|c) - w \nabla_z \log \tilde p(z|\emptyset) \\
  &\propto \nabla_{z} \log\left(
         \tilde p(z|c)^{1+w}\tilde p(z|\emptyset)^{-w}
 \right),
\end{align*}
\ie, $\tilde \scoreest$ estimates the log of the density 
         $\pcfg (z|c) \propto \tilde p(z|c)^{1+w}\tilde p(z|\emptyset)^{-w}$
(see Fig.~\ref{fig:cfg_highlevel}).
Thus, we want to adapt our models to sample from $\pcfg$.
We follow~\cite{ho2022classifier} and train GIVT with an additional null class $\emptyset$.
During inference, we evaluate GIVT twice at every step, once conditional on the actual label $c$ and once conditional on $\emptyset$.
To implement classifier-free guidance, we then have to sample from an unnormalized version of $\pcfg(z)$ derived from the two GIVT predictions.
To this end, we turn to rejection sampling, which requires:
1) an unnormalized density;
2) a good proposal distribution $p'$, that is close to the true target distribution; and
3) a scaling factor $K$ to bound the likelihood ratio between $p'$ and the unnormalized target density.

The distributions we mix are GMMs and finding a good proposal distribution can be challenging. Instead, we first sample the mixture index from $\tilde p(z|c)$ and apply DB-CFG to the corresponding mixture components from $\tilde p(z|c)$ and $\tilde p(z)$ (the components are multivariate Gaussians with diagonal covariance). 
We find empirically that the unconditional components (\ie, distributions predicted using the $\emptyset$ label) tend to have larger variance than the conditional ones (as visualized in Fig.~\ref{fig:cfg_highlevel}).
It is thus sensible to pick sample proposals from $\mathcal N(\mu_c, 2\sigma_c)$, where $\mu_c, \sigma_c$ are the parameters predicted by GIVT when given the label $c$. 
We empirically find that drawing 1000 samples is enough to find at least one valid sample 99.9\% of the time.
For the remaining ${<}0.1$\%, fall back to sampling from $\mathcal N(\mu_c, \sigma_c)$.

We emphasize that the overhead of DB-CFG is small: it requires two forward passes (per inference step) instead of one to predict the conditional and unconditional distribution. We then draw 1000 samples from those in parallel on an accelerator, which is very fast.
We refer to App.~\ref{sec:dbcfg} for Python code.

\begin{figure*}[t]
\centering
\includegraphics[height=3.5cm]{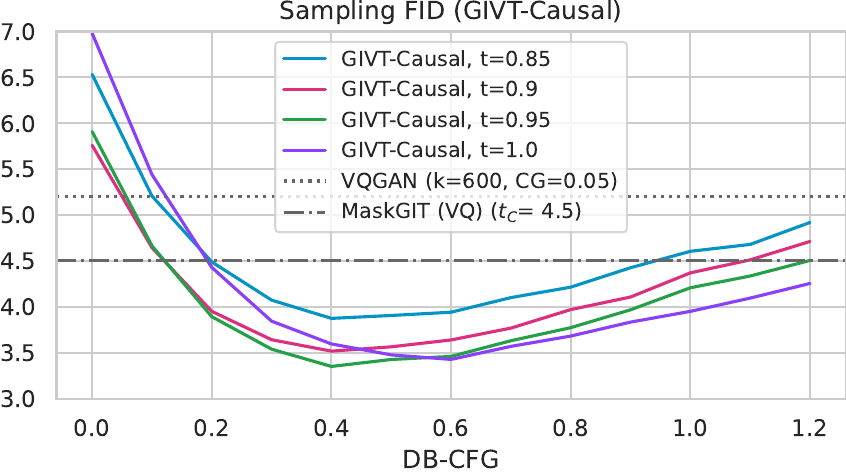}\quad\,
\includegraphics[height=3.5cm]{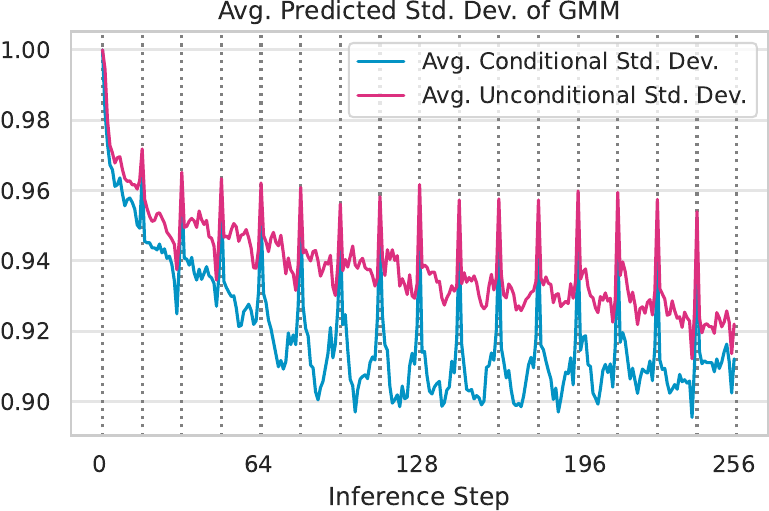}
\caption{\label{fig:cfg_fid_and_scales}
\emph{Left:}
Impact of DB-CFG (Sec.~\ref{sec:dbcfg}) and variance scaling (Sec.~\ref{sec:method:inference:techniques}) on sampling FID 
of our class-conditional $256\times256$ GIVT-Causal models. DB-CFG values in $[0.3, 0.8]$ and variance scaling parameter $t$ in $[0.9, 1.0]$ lead to low FID.
\emph{Right:}
     Average standard deviation of the GMM predicted by GIVT-Causal for class 130, averaged over 128 samples: conditional predictions have lower standard deviation; spikes can be observed when the line changes in the raster scan over the latent feature vectors.
}
\end{figure*}

\section{Experiments} \label{sec:experiments}

\subsection{Image generation}

We use ImageNet1k~\cite{russakovsky2015imagenet} and explore class-conditional generation (where we condition our GIVT on class labels) for 256px and 512px, and \emph{un}conditional generation for 256px.

\fakeparagraph{$\beta$-VAE} We closely follow the setup of MaskGIT \cite{chang2022maskgit}. We use their VAE architecture,
built of ResBlocks (as detailed in App.~\ref{sec:app:trainingdetails}; encoder and decoder have a combined 53.5M parameters), remove the VQ layer and related losses, and replace it with a linear layer predicting $\mu,\sigma$ (Sec.~\ref{sec:method:vae}).
We use the same weights for reconstruction, perceptual, and GAN-loss, as well as identical optimizer parameters, as in \cite{chang2022maskgit, mentzer2023finite}; we only vary the latent dimension $d$ and weight $\beta$ of the KL-term. By default, we set the token dimension to $d=16$ (\ie, the VAE predicts 16 means and variances per token) and $\beta = 5 \cdot 10^{-5}$. We note that our VAE is trained on $256 \times 256$ images, and we also use it for our $512 \times 512$ experiments without retraining (like~\cite{chang2022maskgit}).

\fakeparagraph{GIVT} For GIVT-Causal, we follow the original transformer decoder architecture \cite{vaswani2017attention} in decoder-only mode, but remove biases from attention layers, MLP blocks, and LayerNorms, and replace ReLU by GELU as is common practice. For GIVT-MaskGIT, we simply remove the attention mask during training and feed masked inputs instead of shifted ones. We use the BERT-Large configuration \cite{devlin2018bert} by default (304M parameters), and also explore a larger backbone with 1.67B parameters, denoted with the suffix ``-L'' (see App.~\ref{sec:architecture_details} for details). For model variants with adapter (suffix ``+A''), we use a stack of 8 bijective iRevNet blocks~\cite{jacobsen2018revnet} (with hidden channel dimension $4d$, resulting in 112k additional parameters for $d=16$), applied to the $w \times h \times d$ representation before reshaping it into a sequence.
We configure our GIVT models to predict a $16$-mixture GMM with factorized components (i.e. the mixture components are multivariate Gaussians with diagonal covariance), and explore predicting a single, multivariate Gaussian modeling the full covariance matrix of the tokens as an alternative.
For the conditional generation experiments, we use a learned embedding which we prepend to the embedded token sequence.
To train GIVT, we use Adam with a cosine schedule (500 epochs; with linear warmup of 50 epochs), set the learning rate and weight decay to $10^{-3}$ and $10^{-4}$, respectively, the optimizer $\beta_2$ parameter to $0.95$, the dropout probability to $0.2$ for GIVT-causal and $0.4$ for GIVT-MaskGIT, and the batch size to $8192$.
We use the same data augmentation as during VAE training (see \cite{chang2022maskgit, mentzer2023finite}), and sample from the VAE encoder distribution for every batch (an additional source of randomness besides data augmentation).

We implement GIVT in JAX~\cite{jax2018github} and use distrax \cite{deepmind2020jax} to implement the cand compute the log-probabilities.

\fakeparagraph{GIVT-MaskGIT inference}
Following \cite{chang2022maskgit}, we fix the number of inference steps to 16 and employ the cosine schedule (\ie letting $r=i/16$ at step $i$, the fraction of masked tokens is given by $\cos(\pi/2 r)$). We also sort tokens by likelihood at each step and sample using a ``choice temperature'' $t_C$.

\fakeparagraph{Exploring the VAE latent space} To better understand the interplay between the feature dimension $d$, the KL regularization $\beta$, the reconstruction quality of the VAE, and the sampling quality of the GIVT, we train VAEs with latent dimension in $\{4, 8, 16, 32\}$ and $\beta$ in $\{2.5 \cdot 10^{-5}, 5 \cdot 10^{-5}, 10^{-4}, 2 \cdot 10^{-4}\}$ using the VAE-training setup described at the beginning of this section.
For each of the resulting VAEs we train a GIVT-Causal with the smaller BERT-Base \cite{devlin2018bert} dimensions and a range of values for the number of mixtures $k$.

\fakeparagraph{Evaluation} For the VAEs we report ``reconstruction FID'', the FID obtained when reconstructing the 50k ImageNet validation images.
For our GIVT variants and baselines, we report the sampling FID \cite{heusel2017gans} when sampling a balanced set of 50k images covering all ImageNet classes. In both cases, we rely on the well-established ADM TensorFlow Suite~\cite{dhariwal2021diffusion}, which  uses the entire ImageNet training set as a reference.
Furthermore, we also report Precision and Recall \cite{sajjadi2018assessing}. Finally, we evaluate the representation learning capabilities by training a linear classifier on an average-pooled intermediate representation of  unconditional GIVT-Causal as in prior work~\cite{chen2020generative, yu2021vector} (see App.~\ref{sec:app:probing} for details).

\subsection{Panoptic segmentation and depth estimation}

We build on the UViM framework \cite{kolesnikov2022uvim}, which uses a VQ-VAE to compress the label space of computer-vision dense-prediction tasks, and an encoder-decoder transformer taking the RGB image as an input and predicting the associated dense labels as discrete codes in the VQ-VAE latent space. Here, we replace the VQ-VAE with a $\beta$-VAE and use a GIVT encoder-decoder to model the continuous latent code.
For the VAE, we use the same transformer-based autoencoder architecture (6-layer encoder and 12-layer decoder) and cross-entropy loss as \cite{kolesnikov2022uvim}.
We set $d=16$, $k=1$, and KL weight $\beta=2.5 \cdot 10^{-4}$ for panoptic segmentation and $\beta=2 \cdot 10^{-4}$ for depth estimation.
To build an encoder-decoder GIVT model the same way as in~\cite{kolesnikov2022uvim},
we employ the causal variant described for ImageNet generation and insert a cross-attention layer after each self-attention layer. Following~\cite{kolesnikov2022uvim}, we use the ImageNet-21k-pretrained ViT-L/16 from~\cite{steiner2021train} as the encoder, set the image resolution to 512px, and adopt the preprocessing and optimization hyper-parameters from~\cite{kolesnikov2022uvim}.
We use the UViM variant without encoder and decoder context \cite{kolesnikov2022uvim}.
Finally, we consider variance scaling and beam search, selecting the parameters on a held-out subset of the training set as~\cite{kolesnikov2022uvim}.

\section{Results} \label{sec:results}

\begin{table*}[t]
    \caption{\label{tab:condinet512}
    Results on class-conditional $512{\times}512$ ImageNet.
    We use the standard ADM evaluation suite for metrics, where FID is calculated w.r.t.\ the training set.
    GIVT-MaskGIT obtains competitive FID scores with only 16 inference steps and outperforms its VQ-counterpart. GIVT-Causal-L+A outperforms the best DiT variant, DiT-XL/2-G.
  $^\dagger$Values obtained by us from public code. $^\star$Inference uses activation caching.\vspace{-0.1cm}
    }
    \centering\scriptsize
    \begin{tabular}{llc@{\hskip 1pt}n{2}{2}n{1}{2}n{1}{2}}
        \toprule
        Model 
            & Inference & Steps
            & \multicolumn{1}{c}{FID$\downarrow$}     
            & \multicolumn{1}{c}{Precision$\uparrow$}
            & \multicolumn{1}{c}{Recall$\uparrow$} \\
        \midrule
        ADM~\cite{dhariwal2021diffusion}
                                   &          & 250 & 23.2 & 0.73 & 0.60     \\
        ADM-G~\cite{dhariwal2021diffusion}
                                   & CG $=1.0$  & 250 & 7.72   & {\npboldmath} 0.87 & 0.42   \\
        DiT-XL/2~\cite{peebles2022scalable}
                                   &          & 250 & 12.03 & 0.75 & {\npboldmath} 0.64\\
        DiT-XL/2-G~\cite{peebles2022scalable}
                                   & CFG $=1.5$  & 250 & {\npboldmath}3.04 & 0.84 & 0.54     \\
        \midrule
        MaskGIT~\cite{chang2022maskgit}
            & $t_C=4.5$  & 16  & 7.8007$^\dagger$ & 0.86$^\dagger$ & {\npboldmath} 0.46$^\dagger$ \\
        GIVT-MaskGIT \emph{(Ours)}
            & $t_C=140$ & 16  & {\npboldmath} 4.8635 & {\npboldmath} 0.8785 & {\npboldmath} 0.4770  \\
        \midrule
        GIVT-Causal-L \emph{(Ours)}
            & $t=0.9$ & 512$^\star$ & 8.3493 & 0.7860 & {\npboldmath} 0.6060 \\
        GIVT-Causal-L+A \emph{(Ours)}
            & $t=0.9$, DB-CFG $=0.9$ & 512$^\star$ & {\npboldmath} 2.9163 & {\npboldmath} 0.8429 & 0.5480 \\
        \bottomrule
    \end{tabular} \vspace{-0.1cm}
\end{table*}

\subsection{Image generation} \label{sec:results:image_generation}
\fakeparagraph{VAE latent space}
In Fig.~\ref{fig:kl_abl} we show how varying the weight $\beta$ of the KL term affects 1) the VAE reconstruction FID and 2) the sampling FID of a Base-size GIVT-Causal trained on the corresponding latent sequence. For 1), increasing $\beta$ leads to worse reconstruction FID since the VAE can store less information in the latent. It shifts more of the modeling effort to the VAE decoder, so that the decoder becomes gradually more generative, which affects sampling quality (see \cite[Sec.~7]{tschannen2018recent}\cite{rombach2021high} for more discussion). 

For 2), we see the opposite trend: increasing $\beta$ leads to decreased (better) sampling FID for GIVT models trained on the latent sequence.
Arguably, this is because the VAE latent sequence more closely follows the Gaussian prior, and hence becomes easier for the GIVT to model.
Finally, increasing the number of mixtures $k$ initially reduces the sampling FID substantially, reaching a plateau at $k=16$. We hence set $k=16$ and $\beta=5 \cdot 10^{-5}$ by default, and use a VAE with $\beta=10^{-5}$ for the larger (-L) GIVT models. We emphasize that it is common in the literature to explore and tune hyper-parameters such as $\beta$ or analogously the vocab size and commitment loss in VQ \cite[Table~5]{esser2020taming} \cite[Table~8]{rombach2021high}\cite{peebles2022scalable}\cite[Fig.~3]{mentzer2023finite}.

\fakeparagraph{Sampling FID}
In Table~\ref{tab:fid_results} we show the sampling FID for four model classes on class-conditional $256\times256$ ImageNet:
GANs, diffusion-based approaches, as well as masked and sequence modeling approaches. GIVT-MaskGIT outperforms MaskGIT~\cite{chang2022maskgit} which has comparable model size and inference cost, and DB-CFG leads to an additional improvement. In absence of guidance techniques, our GIVT-Causal models outperform all diffusion baselines as well as VQGAN by a large margin. Using guidance techniques, GIVT-Causal obtains FID of 3.35 compared to 5.20 for VQGAN with a more than $4.5\times$ smaller model (0.3B for GIVT vs.\ 1.4B parameters), and also outperforms the 32\% larger LDM-4-G.
Our larger GIVT-Causal-L+A obtains 16\% and 17\% reduction in FID without and with guidance, respectively, compared to ViT-VQGAN which has the same generative transformer size but a $4\times$ larger sequence length (resulting in more than $4\times$ slower sampling) and a $10\times$ larger VAE.

We present sampling FID for $512\times512$ ImageNet in Table~\ref{tab:condinet512}. GIVT-MaskGIT obtains a 38\% lower FID than MaskGIT with comparable model size and inference cost. GIVT-Causal-L+A outperforms DiT-XL/2, the best available DiT model, both without and with guidance (albeit with a larger model).

Finally, we present \emph{un}conditional results in App.~\ref{sec:app:uncond}. This task is considerably harder, but GIVT-Causal beats the diffusion-based ADM~\cite{dhariwal2021diffusion} by a large margin.

\begin{table}[t]
    \caption{UViM based on GIVT-Causal and VQ-VAE evaluated on panoptic segmentation (COCO Panoptic 2017) and depth estimation (NYU Depth v2). We report the panoptic quality (PQ) and RMSE for the VAE/VQ-VAE reconstructions of the validation set label maps (recon.) and the inference metrics on the actual dense prediction tasks (inference). GIVT obtains metrics comparable to the VQ-based UViM.}
    \centering
    \scriptsize
    \vspace{-0.1cm}
    \begin{tabular}{@{\hskip 0pt}lcccc@{\hskip 0pt}}
    \toprule
    &  \multicolumn{2}{c}{COCO Pan. (PQ$\uparrow$)} & \multicolumn{2}{c}{NYU Depth v2 (RMSE$\downarrow$)} \\
    \cmidrule[0.5pt](l{2pt}r{2pt}){2-3} \cmidrule[0.5pt](l{2pt}r{2pt}){4-5}
    & recon. & inference & \quad recon. & inference \\ \midrule
       UViM \cite{kolesnikov2022uvim} & \color{gray} 66.0 & 39.0 &  \quad \color{gray} 0.183 & {\bf 0.459} \\
       GIVT (ours) & \color{gray} 71.0 & {\bf 40.2} & \quad \color{gray} 0.195 & 0.474  \\ \bottomrule
    \end{tabular}
    \label{tab:coco-panoptic} \vspace{-0.1cm}
\end{table}

\fakeparagraph{Ablations and visualizations} 
Fig.~\ref{fig:sampling_alg_abl} compares the effect of model configuration (number of mixtures $k$, adapter) and sampling algorithm (variance scaling, beam search, DB-CFG) on FID. For every model configuration, all the sampling algorithms lead to solid improvements in FID, with DB-CFG being the most effective one across all configurations. Increasing $k$ from 1 to 16 overall leads to somewhat larger improvements than keeping $k=1$ and adding an adapter. Moreover, combining adapter with $k=16$ results in compounding improvements across sampling algorithms. 

Fig.~\ref{fig:cfg_fid_and_scales} (left) shows the impact of the variance scaling and CFG parameters on the sampling FID. In Fig.~\ref{fig:cfg_fid_and_scales} (right), we visualize the predicted standard deviation as a function of the GIVT-Causal inference step. The standard deviation gradually decreases, meaning that the predictions later in the sampling process become more certain. Furthermore, the unconditional predictions generally have a higher standard deviation, as expected.

For GIVT-MaskGIT, predicting a single Gaussian with full covariance matrix per latent vector, rather than assuming a diagonal covariance, only led to very modest gains of about 3\%. A GMM with factorized component densities therefore seems to be the more effective alternative. Furthermore, a full covariance matrix makes DB-CFG less tractable than the diagonal covariance (because the high dimensional multivariate distribution has more regions of low density).

\fakeparagraph{Samples}
Fig.~\ref{fig:ar_sampling} shows ten $512\times512$ samples from GIVT-Causal-L+A,
and App.~\ref{sec:app:samples} shows samples for other GIVT-Causal variants and GIVT-MaskGIT.
We can see that the model produces high-fidelity, coherent samples. 
To study sample diversity, we show multiple samples from different models for a fixed class.

In Fig.~\ref{fig:maskgit_sampling} in App.~\ref{sec:app:samples}, one can see two samples from our VAE (obtained by decoding latents sampled from the prior), which show a soup of image textures. We then show different steps of the GIVT-MaskGIT inference, and observe similar behavior as in the VQ-based model (\cite[Fig.~2]{chang2022maskgit}).

\begin{wraptable}{r}{6.7cm}
\vspace{-1.2cm}
    \caption{\label{tab:linear_imagenet} ImageNet linear probing accuracy of unconditional GIVT-Causal and generative models from the literature. GIVT-Causal matches VIM+ViT (ViT-VQ-GAN)~\cite{yu2021vector} which has more than $2\times$ the model parameters and $4\times$ the sequence length (and hence FLOPs). \emph{Type}: (Latent) generative model type. \emph{\#Param.}: Number of parameters of the full (latent) generative model.
    }
    \centering\scriptsize
    \begin{tabular}{llrr@{\hskip 1pt}r}  % 
        \toprule
        Model & Type & \#Tok. & \#Param.
            & \multicolumn{1}{c}{Acc.$\uparrow$}     \\
        \midrule
        BigBiGAN~\cite{donahue2019large} & & & 344M & 61.3 \\
        iGPT-L~\cite{chen2020generative} & dec.-only & 1024 &  1362M & 60.3 \\
        VIM+CNN~\cite{yu2021vector} & dec.-only & 1024 & 650M & 61.8 \\
        VIM+ViT~\cite{yu2021vector} & dec.-only & 1024 & 650M & {\bf 65.1} \\
        \color{gray} MAGE ViT-L \cite{li2023mage} & \color{gray} enc.-dec. & \color{gray} 256 & \color{gray} 404M & \color{gray} 78.9 \\
        \midrule
        GIVT-Causal \emph{(Ours)} & dec.-only & 256 & 304M & {\bf 65.1} \\
        \bottomrule
    \end{tabular}
\vspace{-0.5cm}
\end{wraptable}

\fakeparagraph{Representation learning} Table~\ref{tab:linear_imagenet} shows the ImageNet linear probing accuracy of unconditional GIVT-Causal and generative models from the literature (we chose the model variants closest in terms of model size and compute). GIVT-Causal matches VIM+ViT (ViT-VQGAN)~\cite{yu2021vector} which has more than $2\times$ the model parameters and $4\times$ the sequence length (and hence FLOPs). GIVT-Causal is only outperformed by MAGE~\cite{li2023mage}, whose latent encoder-decoder architecture is better suited for representation learning than decoder-only models. An investigation of the probing accuracy as function of the layer index can be found in App.~\ref{sec:app:probing}.

\subsection{Panoptic segmentation and depth estimation}

Table~\ref{tab:coco-panoptic} compares the performance of a GIVT-based UViM variant with a VQ-VAE-based baseline (both without encoder/decoder context) on COCO Panoptic 2017~\cite{kirillov2019panoptic} and NYU Depth v2~\cite{silberman2012indoor}. We report the panoptic quality metric (PQ)~\cite{kirillov2019panoptic} and RMSE, respectively, and find that
our GIVT-based model outperforms the baseline in panoptic segmentation and performs slightly worse in depth estimation.
In App.~\ref{sec:app:samples} we show visual results.

\section{Conclusion} \label{sec:conclusion}

In this paper, we proposed simple modifications to standard transformer decoder-only models enabling them to generating real-valued vectors. To our knowledge, this is the first decoder-only model amenable to generating sequences of real-valued vectors. In the context of image generation with VQ-GAN or Mask-GIT, this side-steps training difficulties such as low codebook usage in VQ-VAEs and corresponding mitigations like entropy losses or codebook-splitting algorithms, by enabling the use of standard VAEs which are much easier to train. Furthermore, our method avoids large embedding matrices because the feature representations can directly be consumed and predicted by our GIVT model. Our simple, quantization-free approach outperforms its VQ-based counterparts in class-conditional image generation and image representation learning, often by significant margins. GIVT also obtains strong performance in dense prediction tasks when applied to UViM. We hope that future work explores applications of GIVT to other modalities such as audio and time-series modeling.

\section*{Acknowledgments}

We would like to thank André Susano Pinto, Neil Houlsby, Eirikur Agustsson, Lucas Theis, and Basil Mustafa for inspiring discussions and helpful feedback on this project. We also thank Han Zhang for support with the VAE training code.

\bibliographystyle{splncs04}
\bibliography{main}

\clearpage

\appendix
% \pagestyle{plain}
% \setcounter{page}{1}
% \section*{GIVT: Generative Infinite-Vocabulary Transformers --- Supplementary Material}

\definecolor{codegreen}{rgb}{0,0.6,0}
\definecolor{codegray}{rgb}{0.5,0.5,0.5}
\definecolor{codepurple}{rgb}{0.58,0,0.82}

\lstdefinestyle{mystyle}{
    % backgroundcolor=\color{backcolour},   
    commentstyle=\color{codegreen},
    keywordstyle=\color{magenta},
    numberstyle=\tiny\color{codegray},
    stringstyle=\color{codepurple},
    basicstyle=\ttfamily\small,
    breakatwhitespace=false,         
    breaklines=true,                 
    captionpos=b,                    
    keepspaces=true,                 
    % numbers=left,                    
    % numbersep=5pt,                  
    showspaces=false,                
    showstringspaces=false,
    showtabs=false,                  
    tabsize=2
}

\section*{arXiv version history}

\begin{itemize}
    \item v1: Initial version.
    \item v2: Adds details on loss.
    \item v3: Multiple model updates:
    \begin{itemize}
        \item GMM: Changes from $d$ per-channel scalar GMMs or a single Gaussian prediction to a $d$-variate GMM with factorized components (\ie, component distributions are $d$-variate Gaussians with diagonal covariance).
        \item Introduces adapters.
        \item Large GIVT-Causal models (1.67B parameters) and GIVT-Causal results for 512px image generation.
        \item Changes optimizer from AdaFactor to Adam due to model sharding (this change is performance-neutral).
        \item Updates image generation results with these new model variants.
    \end{itemize}
    \item v4: ECCV 2024 camera ready version (minor changes).
\end{itemize}

\section{Additional discussions}

\fakeparagraph{Why continuous latents for images?}  \label{sec:additional_discussion}
Continuous latents naturally fit intrinsically continuous-valued data such as image representations, and avoid excessive compression and the resulting information loss induced by VQ. Removing VQ also avoids the well-documented challenges of stochastic-gradient-based discrete representation learning: VQ-VAE~\cite{van2017neural} and its variations require a VQ optimization problem, which on its own is NP-hard, to be solved jointly with an embedding learning problem. A large body of literature~\cite{chang2022maskgit, huh2023straightening, kolesnikov2022uvim, lancucki2020robust, van2017neural} focuses on mitigating issues resulting from these optimization difficulties, such as low vocabulary utilization (Sec.~1). For example, [46] avoids vector quantization by using a product codebook, but still relies on non-differentiable scalar quantization. In contrast, with GIVT we obtain SOTA results based on a $\beta$-VAE which does not include any non-differentiable operations and does not require any advanced tricks from the VQ literature~\cite{chang2022maskgit, huh2023straightening, kolesnikov2022uvim, lancucki2020robust, van2017neural}. 

\fakeparagraph{Continuous latents vs. infinite vocabulary} In theory, continuous (\ie, real-valued) latents/tokens always imply an infinite vocabulary when mapping them \emph{exactly} to a discrete code. In other words, the infinite vocabulary is a consequence of continuous latents. In practice, continuous latents are represented with finite precision, \eg, \texttt{float32}, but using the raw 32-bit representation as a discrete code would still imply an impractically large vocabulary size $2^{32} \approx 4.3$B for a single latent dimension. GIVT directly models the continuous latents and avoids materializing this vocabulary.

\section{Architecture details} \label{sec:architecture_details}

Architecture details for different model variants used for image generation are listed in Table~\ref{tab:arch_details_app}. We rely on the standard pre-LN setup (see, \eg, UViM \href{https://github.com/google-research/big_vision/blob/main/big_vision/models/proj/uvim/vtt.py}{vtt.py} for a concrete implementation). For the UViM experiments, we use GIVT-Causal with the Default config and a cross-attention layer inserted after every self-attention layers as in~\cite{kolesnikov2022uvim} to fuse visual features extracted from the input RGB image.

Adapters are constructed by stacking 8 convolutional, bijective iRevNet blocks \cite{jacobsen2018revnet} with hidden channel dimension $4d$ (resulting in 112k additional parameters for $d=16$). Each block consists of 3 Conv layers, interleaved with GroupNorm and ReLU, and has identical input and output shapes $w \times h \times d$ (\ie, the adapter is applied to the VAE latent $z$ before reshaping it into a sequence). We base our iRevNet block on the reference implementation \href{https://github.com/jhjacobsen/pytorch-i-revnet/blob/master/models/iRevNet.py}{iRevNet.py}, replacing BatchNorm with GroupNorm and removing subsampling layers.

\begin{table*}[h!]
\hspace{-0.2cm}
    \caption{\label{tab:arch_details_app}Architecture details for different model variants for image generation. We also specify the KL weight $\beta$ used to train the associated VAE. The Base architecture is used to explore the feature dimension $d$, number of mixtures $k$ and $\beta$ (Fig.~\ref{fig:sampling_alg_abl}). We use $\beta=5\cdot10^{-5}$ by default, and $\beta=10^{-5}$ for Large models. $^*$GIVT-Causal-L at 512px relies on a space-to-depth transformation, where pairs of consecutive 16-dimensional feature vectors are stacked into 32-dimensional vectors, resulting in sequence length of 512 instead of 1024.}
    \centering\scriptsize
    \setlength{\tabcolsep}{1pt}
    \begin{tabular}{llcccrcccrrcc}
        \toprule
        Model & Size & Res. & $d$ & $k$ & Width & Depth & MLP & Heads & Param. & Tok. & Drop. & $\beta$ \\
        \midrule
        Causal & Base & 256 & $4\ldots32$ & $1\ldots32$ & 768 & 12 & 3072 & 12 & 86M & 256 & 0.1 & $0.25\ldots2\cdot10^{-4}$\\
        Causal & Default & 256 & 16 & 16 & 1024 & 24 & 4096 & 16 & 304M & 256 & 0.2 & $5\cdot10^{-5}$\\
        Causal & Large & 256 & 16 & 16 & 1536 & 48 & 8192 & 16 & 1.67B & 256 & 0.3 & $10^{-5}$\\
        Causal$^*$ & Large & 512 & 32 & 32 & 1536 & 48 & 8192 & 16 & 1.67B & 512 & 0.1 & $10^{-5}$\\
        \midrule
        MaskGIT & Default & 256 & 16 & 16 & 1024 & 24 & 4096 & 16 & 304M & 256 & 0.4 & $5\cdot10^{-5}$\\
        MaskGIT & Default & 512 & 16 & 16 & 1024 & 24 & 4096 & 16 & 304M & 1024 & 0.4 & $5\cdot10^{-5}$\\
        \bottomrule
    \end{tabular}
\end{table*}

\begin{figure*}[h!]
    \centering
    \includegraphics[width=0.8\linewidth]{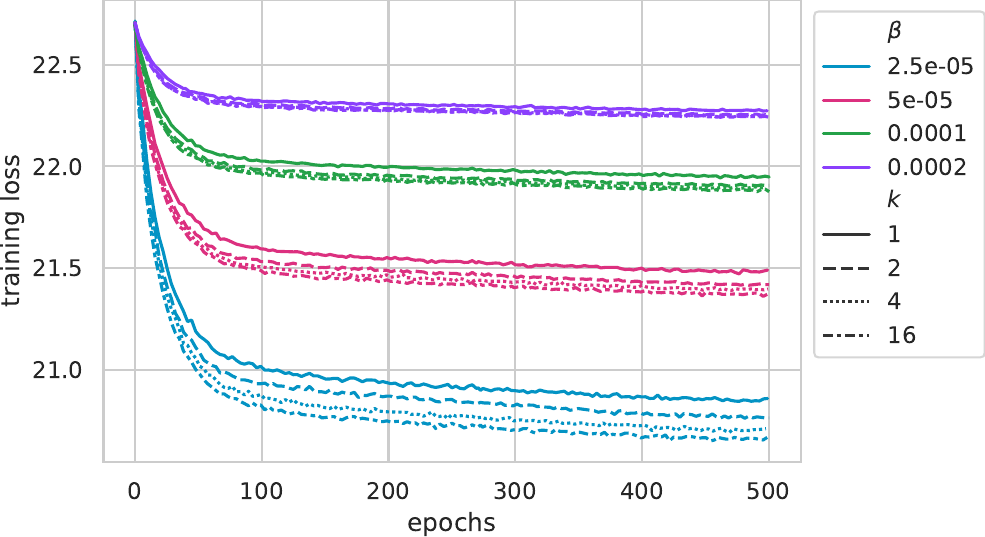}
    \caption{\label{fig:training_loss} Loss (NLL) curves when training the Base-sized GIVT-causal models from Fig.~\ref{fig:sampling_alg_abl}, with latent dimension $d=16$, as a function of the number of mixtures $k$ and the KL weight $\beta$ in the $\beta$-VAE. Reducing $\beta$ or increasing $k$ leads to a reduction in loss, which not always translates into a lower FID. The loss is computed using distrax as described in Sec.~\ref{app:loss_function}. Alternative implementations can lead to scaling by $\frac{1}{d}$ ($\frac{1}{16}$ here).}
\end{figure*}

\section{Training details} \label{sec:app:trainingdetails}

\subsection{Loss function and implementation details} \label{app:loss_function}

We start with the case where GIVT models the distribution of each feature vector or soft token in the sequence, conditionally on all previous feature vectors in the sequence, with a $k$-mixture GMM whose components are independent across channels (\ie, the components are multivariate Gaussian distributions with diagonal covariance matrix). The predicted GIVT output for a mini batch of size $B$ of $d$-dimensional feature sequences of length $L$ is a $B \times L \times (2dk + k)$ tensor $y = [m; s; \pi]$ where $m, s$ are $B \times L \times dk$ tensors and $\pi$ is a $B \times L \times k$ tensor, all stacked along the last dimension, comprising means, standard deviations, and mixture weights, respectively. We assume the entries of $s$ to be lower-bounded by a small positive constant $\epsilon = 10^{-5}$ (\eg, by applying a softplus function to the corresponding network outputs and clipping to $\epsilon$) and the entries of $\pi$ to be non-negative. $m$ is composed as $[m^{(1)}; \ldots; m^{(k)}]$ where the $m^{(n)}$ are the $B \times L \times d$ mean tensors of the GMM components stacked along the last axis; $s$ is composed in the same way. The mixture weights are assumed normalized across components (\eg, via a softmax function), \ie, $\sum_{n=1}^k \pi_{b,\ell}^{(n)} = 1$ for all $b,\ell$.

Then, the negative log-likelihood (NLL) of the feature sequences $\{z_b\}_{b=1}^B$ (\ie the training images after embedding via the VAE encoder and applying reparametrization) w.r.t. the distribution $\tilde p$ predicted by GIVT can be written as
\begin{align}
    &-\sum_{b=1}^B \log(\tilde p(z_b)) \nonumber \\
    &=-\sum_{b=1}^B \log\left(\prod_{\ell = 1}^L \left(\sum_{n=1}^k \pi_{b, \ell}^{(n)} \prod_{c = 1}^d \mathcal N(z_{b, \ell, c} | m_{b, \ell, c}^{(n)}, s_{b, \ell, c}^{(n)})\right)\right) \nonumber \\
    &=-\sum_{b=1}^B \sum_{\ell = 1}^L \log\left(\sum_{n=1}^k \pi_{b, \ell}^{(n)} \prod_{c = 1}^d \mathcal N(z_{b, \ell, c} | m_{b, \ell, c}^{(n)}, s_{b, \ell, c}^{(n)})\right), \label{eq:gmmll}
\end{align}
where $\mathcal N(z | m, s)$ is a Gaussian density
\begin{equation*}
    \mathcal N(z | m, s) = \frac{1}{s \sqrt{2 \pi}} e^{-\frac12 \left(\frac{z - m}{s}\right)^2}.
\end{equation*}
When using a single Gaussian instead of a GMM to model each output channel (\ie $k=1$ s.t. $\pi_{b, \ell}^{(1)} = 1$ for all $b, \ell$) Eq.~\eqref{eq:gmmll} becomes
\begin{equation*}
    \sum_{b=1}^B \sum_{\ell = 1}^L \sum_{c = 1}^d \frac12 \left(\frac{z_{b, \ell, c} - m_{b, \ell, c}}{s_{b, \ell, c}}\right)^2 + \log(s_{b, \ell, c}) + \frac12 \log (2\pi).
\end{equation*}
While this loss simplifies to a sum over the sequence dimension $\ell$, note that $m_{b, \ell, c}, s_{b, \ell, c}$ (and in the GMM case $\pi_{b, \ell, c}$) are predicted from $z_{b,1},\ldots, z_{b,\ell-1}$ by GIVT (setting $z_{b,0}$ to a learned \texttt{[CLS]} or \texttt{[BOS]} vector). Further, it can be seen that lower-bounding the $s_{b, \ell, c}$ with $\epsilon > 0$ avoids extremely large loss values when some $z_{b,\ell}$ fall into low-density regions of $\tilde p$.

Fig.~\ref{fig:training_loss} shows the training loss curves as a function of VAEs with different KL weights $\beta$ and $k$, following the setup of Fig.~\ref{fig:kl_abl}. Increasing $k$ and reducing $\beta$ leads to lower loss values. Note, however, that a lower loss does not always lead to a lower sampling FID (see Fig.~\ref{fig:kl_abl}). Importantly, initial and final loss values, as well as the relative reduction in training loss throughout training can differ significantly from the values typically observed for the discrete cross-entropy or NLL, \eg, in language modeling.

For the multivariate case, where dependencies between feature channels are modeled, GIVT predicts a $d \times d$ covariance matrix $S^{(n)}_{b,\ell}$ for each mixture component $n$, and data negative log-likelihood becomes
\begin{equation*}
    -\sum_{b=1}^B \sum_{\ell = 1}^L \log\left(\sum_{n=1}^k \pi_{b, \ell}^{(n)} \tilde{ \mathcal N}(z_{b, \ell} | m_{b, \ell}^{(n)}, S_{b, \ell}^{(n)})\right),
\end{equation*}
where $\tilde{\mathcal N}$ is a multivariate Gaussian density.

For all loss variants, the MaskGIT version of GIVT the sum over $L$ is reduced to the sum over the indices $l$ correspond to masked locations, and the loss is normalized by the number of mask tokens.

Finally, loss computation and sampling can easily be implemented with dedicated deep learning packages, for example the JAX library distrax~\cite{deepmind2020jax}:

\begin{lstlisting}[language=Python, style=mystyle, basicstyle=\scriptsize]
# With z and m, s, pi as in Eq. (2)
pdf = distrax.MixtureSameFamily(
    mixture_distribution=distrax.Categorical(logits=pi),
    components_distribution=distrax.MultivariateNormalDiag(
        loc=m, scale_diag=s))
# Sample from next token distribution (teacher forcing)
samples = pdf.sample()
# Compute NLL
loss = -pdf.log_prob(z).mean()

\end{lstlisting}

\subsection{Image generation}

For the image generation experiments on ImageNet, we adapt the CNN-based VQ-GAN tokenizer from MaskGIT (see \href{https://github.com/google-research/maskgit/blob/main/maskgit/nets/vqgan_tokenizer.py}{vqgan\_tokenizer.py}). We replace the VQ layer and related losses with a Gaussian reparametrization layer~\cite{kingma2013auto}, and we use the hyper parameters given in Table~\ref{tab:hp:cnn} for the CNN. See Sec.~\ref{sec:architecture_details} for details on the GIVT model architecture and variants.

\begin{table}[h]
    \caption{\label{tab:hp:cnn}Hyperparameters of the ImageNet CNN-based VAE encoder/decoder. Note that the 32 features of the embedding are split into 16 means and 16 scales, so our actual representation has $d=16$ channels after reparametrization.}
    \centering
    \begin{tabular}{ll}
    \toprule
    \texttt{embedding\_dim} & 32 \\
    \texttt{filters} & 128 \\
    \texttt{num\_res\_blocks} & 2 \\
    \texttt{channel\_multipliers} & (1, 1, 2, 2, 4) \\
    \texttt{conv\_downsample} & False \\
    \texttt{activation\_fn} & ``swish'' \\
    \texttt{norm\_type} & ``GN'' \\
    \bottomrule
    \end{tabular}
\end{table}

\begin{table*}[t]
    \caption{\label{tab:uncondinet}
    Results for \textbf{un}conditional $256 \times 256$ ImageNet generation. We use the standard ADM evaluation suite for metrics, where FID is calculated w.r.t.\ the training set. We obtained samples for MAGE~\cite{li2023mage} using their \href{https://github.com/LTH14/mage}{GitHub code}. The MAGE Large models have considerably more model parameters than our GIVT-Causal models because MAGE has a latent encoder-decoder model (rather than a decoder-only model).
    }
    \centering\small
    \begin{tabular}{lln{2}{2}n{1}{2}n{1}{2}}
        \toprule
        Model 
            & Inference 
            & \multicolumn{1}{r}{FID$\downarrow$}
            & \multicolumn{1}{c}{Precision$\uparrow$}
            & \multicolumn{1}{c}{Recall$\uparrow$} \\
        \midrule
        ADM~\cite{dhariwal2021diffusion}
                                   &          & 26.2 & 0.61 & 0.63  \\
        MAGE (ViT-B)
            & temp $=6.0$ & 12.080213022509724 & 0.61988 & 0.6364 \\
        MAGE (ViT-L)                   
            & temp $=6.0$ & 9.9528908191744 & 0.65928 & 0.6589 \\
        GIVT-Causal \emph{(Ours)}
            & $k=1$, $t=0.9$ & 19.9226 & 0.6259 & 0.6013 \\
        GIVT-Causal \emph{(Ours)}
            & $k=16$, $t=0.95$ & 17.6995 & 0.6587 & 0.6282 \\
        GIVT-Causal-L \emph{(Ours)}
            & $k=16$, $t=0.95$ & 11.0204 & 0.7207 & 0.6015\\  %xid/74643060
        \bottomrule
    \end{tabular}
\end{table*}

\fakeparagraph{ImageNet preprocessing}
We preprocess the ImageNet data as follows, following prior work~\cite{chang2022maskgit, mentzer2023finite}:
\begin{itemize}
    \item Decode JPEGs
    \item Random crop such that between 80\% and 100\% of the source image are retained
    \item Resize to target resolution ($256{\times}256$ or $512{\times}512$) using bicubic filters with anti-aliasing
    \item Randomly flip the image horizontally with probability~0.5
\end{itemize}

\subsection{Panoptic Segmentation and Depth Estimation}

For our UViM panoptic segmentation and depth estimation experiments, we adopt the public \href{https://github.com/google-research/big_vision/blob/main/big_vision/configs/proj/uvim/README.md}{UViM GitHub repo} and only replace the VQ layer in stage I, adapt the transformer decoder in stage II, and modify the losses accordingly.

\section{DB-CFG implementation} \label{sec:app:dbcfg}

We show the JAX implementation of the rejection sampler we use for DB-CFG in Fig.~\ref{fig:dbcfg:impl}.

\section{Unconditional image generation} \label{sec:app:uncond}

In Table~\ref{tab:uncondinet} we present FID results for unconditional ImageNet generation.

\begin{figure}[t]
    \centering
    \includegraphics[width=0.7\linewidth]{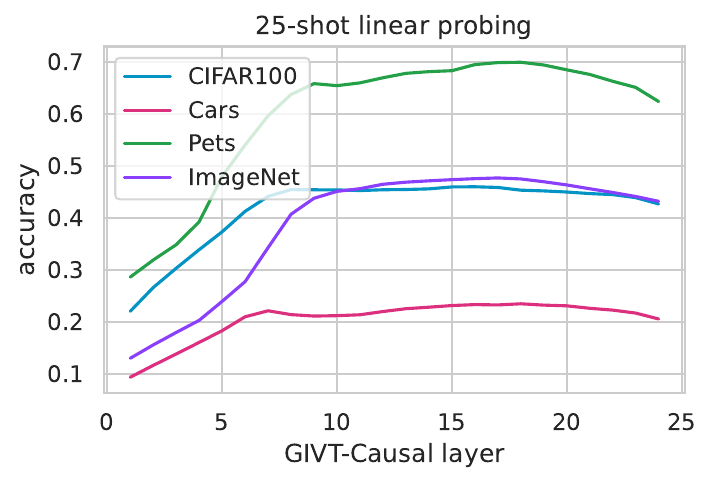}
    \caption{\label{fig:few_shot_probing}25-shot linear probing accuracy as a function of the GIVT-Causal layer trained on unconditional $256 \times 256$ ImageNet. The range of layers 10 to 20 lead to high accuracy for all data sets.}
\end{figure}

\begin{figure*}[t]
    \centering
    \includegraphics[width=0.48\linewidth]{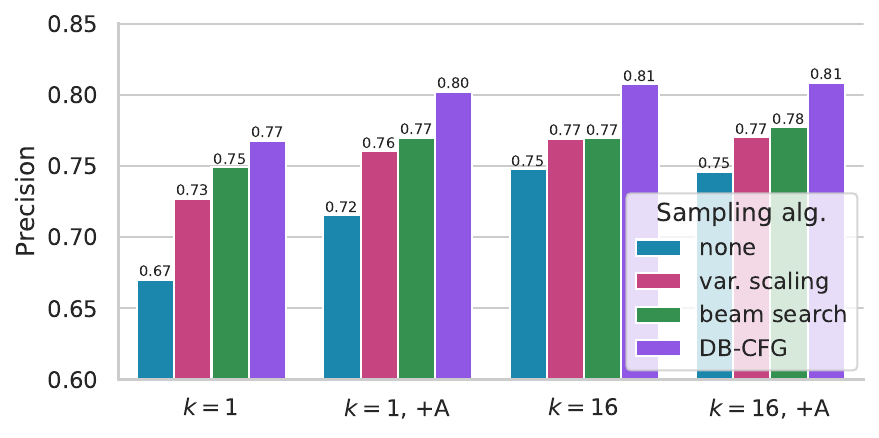} \quad
    \includegraphics[width=0.48\linewidth]{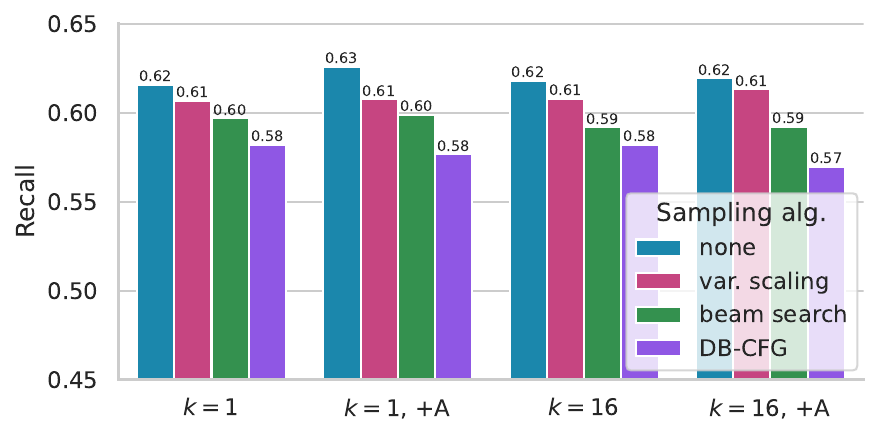}
    \caption{\label{fig:sampling_alg_abl_app} Effect of different sampling strategies and model variants (GIVT-Causal-L) on precision and recall, complementing Fig.~\ref{fig:sampling_alg_abl} for FID. Increasing the number of mixtures $k$ and adding an adapter have lead to compounding effects. DB-CFG is the most effective sampling strategy to increase precision for all model variants.
    }
\end{figure*}

\section{Probing intermediate representations} \label{sec:app:probing}

We extract intermediate representations form a GIVT-causal trained for $256 \times 256$ unconditional ImageNet generation by average-pooling the output of a given layer over the sequence dimension (which results in a feature vector of dimension 1024). Thereby, the input image is first encoded with the VAE encoder and then fed to GIVT as during teacher forcing (\ie the latent sequence is first right-shifted and padded, and then fed to GIVT with causal attention mask).

Following prior work~\cite{chen2020generative, yu2021vector}, we investigate the impact of the layer index on the linear probing accuracy. To speed up the evaluation we rely on 25-shot classification with the fast evaluator form~\cite{zhai2022scaling} and test a range of diverse data sets. The results presented in Fig.~\ref{fig:few_shot_probing} show that the range of layers 10 to 20 lead to high accuracy for all data sets.

We choose layer 18 for the linear probing experiments using the full ImageNet training set. To select hyperparameters we set aside 1\% of the training data and consider the selection of schedules and hyperparameters described in~\cite[App.~A]{tschannen2023image}. The results, along with other generative models from the literature, can be found in Table~\ref{tab:linear_imagenet}. See Sec.~\ref{sec:results:image_generation} for a discussion of these results.

\section{Comparison of methods in terms of FLOPs}

Among the models reported in Table~\ref{tab:fid_results} only the two diffusion models~[14, 51] report FLOPs, but not the sequence or masked models, which are most closely related to our GIVT models. Assuming the same autoregressive transformer optimizations (e.g. caching) and VAE inference cost, we obtain the following approximate comparison, using the $N$ FLOPs of GIVT-Causal-L+A as a reference. MaskGIT and GIVT-MaskGIT require virtually the same number of FLOPs for inference.

\begin{table}[h]
\caption{Approximate comparison in FLOPs of GIVT and autoregressive baselines from the literature.}
    \small
    \setlength{\tabcolsep}{2pt}
    \begin{center}
    \begin{tabular}{llrrrr}  % 
        \toprule
        Model & Guidance & Steps & Params. & FLOPs & FID \\
        \midrule
        GIVT-Causal-L+A & -- & 256 & 1.67B & $N$ & 3.46 \\
        VQGAN [20] & -- & 256 & 1.4B & $\simeq N$ & 17.04 \\
        ViT-VQGAN-L [76] & -- & 1024 & 1.7B & $\simeq 4 N$ & 4.17 \\ 
        \midrule
        GIVT-Causal-L+A & DB-CFG & 256 & 1.67B & $\simeq2N$ & 2.59 \\
        VQGAN [20] & CG $=0.05$ & 256 & 1.4B & $\simeq 20 N$ & 5.20 \\
        ViT-VQGAN-L [76] & CG $=0.5$ & 1024 & 1.7B & $\simeq 8 N$ & 3.04 \\ 
        \bottomrule
    \end{tabular}
    \end{center}
\end{table}

\section{Broader impact}

The method described in this paper modifies transformer decoder-only models to enable generation of sequences of feature vectors, which can be seen as a research endeavor on the foundation of transformer models, with many potential downstream applications. The broader impact will depend largely on the downstream application, and how it is affected by this work.

Here, we focus on generation of visual data. For class-conditional image generation, we rely on ImageNet, and the resulting models provide basic control of the generated image content via the class label. These models do not allow for fine-grained control of image content, or manipulation of existing images. The biases and issues inherent with ImageNet are well-studied and documented [Uday Prabhu and Birhane, 2020]. We expect that our image generation models could reflect some of these issues, and we advise users to use and deploy the model carefully, taking these issues into consideration.

It is relatively straight-forward to extend our image generation models to a text-to-image interface, which enables training on much larger and less curated image/text data sets collected from the web. Such models allow for much more fine-grained control and hence could be misused for malicious purposes such as spreading misinformation or identity theft. Furthermore, such models might reflect biases in large non-curated data sets from the web, such as harmful stereotypes. The deployment and release of such models thus has to be handled with even more care.

\lstset{style=mystyle}

\begin{figure*}
\centering
\begin{minipage}[c]{\linewidth}%
\begin{lstlisting}[language=Python, basicstyle=\scriptsize]
import jax
import jax.numpy as jnp
import tensorflow_probability as tfp
tfd = tfp.substrates.jax.distributions

# We assume the ps have shape (b, seq_len, c).
def rejection_sample(
    seed, p_cond: tfd.Normal, p_uncond: tfd.Normal, w, max_samples=1_000
):
  rng_sample, rng_uni = jax.random.split(seed, 2)
  scale_simple = jnp.stack([p_cond.scale, p_uncond.scale], -1).max(-1) * 2
  simple = tfd.Normal(loc=p_cond.loc, scale=scale_simple)  # Proposal. 

  def unnormalized_pcfg(x):
    return jnp.exp((1 + w) * p_cond.log_prob(x) - w * p_uncond.log_prob(x))

  # Find scaling factor by checking for max around the conditional loc.
  points = p_cond.loc[None, ...] + jnp.linspace(-10, 10, 1001).reshape(
      -1, 1, 1, 1)
  fac = jnp.max(unnormalized_pcfg(points) / simple.prob(p_cond.loc), axis=0)

  # Shape (max_samples, b, seq_len, c)
  xs = simple.sample(seed=rng_sample, sample_shape=(max_samples,))
  facq = fac * simple.prob(xs)
  ys = jax.random.uniform(rng_uni, shape=facq.shape, minval=0.0, maxval=facq)
  p = unnormalized_pcfg(xs)
  # Mask is True if the sample is valid. 
  mask = ys < p
  # Now we need to do tricks to get the first element in `mask` that is
  # True, in a jit-able way. We do this by making a shifted mask that is 
  # False for every element after the first True. Example:
  # mask            [0, 1, 0, 1, 0, 0, 1, 0]
  # cmask           [0, 1, 1, 1, 1, 1, 1, 1]
  # shifted_cmask   [0, 0, 1, 1, 1, 1, 1, 1]
  # keep            [0, 1, 0, 0, 0, 0, 0, 0]  # <- picks the first valid!
  cmask = jnp.cumsum(mask, axis=0).astype(jnp.bool_)
  shifted_cmask = jnp.pad(
      cmask, [(1, 0), (0, 0), (0, 0), (0, 0)], constant_values=False
  )[:-1]
  assert shifted_cmask.shape == mask.shape
  keep = jnp.logical_and(cmask, jnp.logical_not(shifted_cmask))
  sample = jnp.where(keep, xs, 0).sum(0)  # Grab the first valid sample.
  ok = mask.sum(0) > 0  # Fall back to pdf_c if not ok.
  return jnp.where(ok, sample, pdf_c.sample(seed=rng_sample))
\end{lstlisting}
\end{minipage}
\caption{\label{fig:dbcfg:impl}\texttt{jax.jit}-compatible implementation of the rejection sampler for DB-CFG.}
\end{figure*}

\begin{figure*}[hb]
    \centering
    \includegraphics[width=\linewidth]{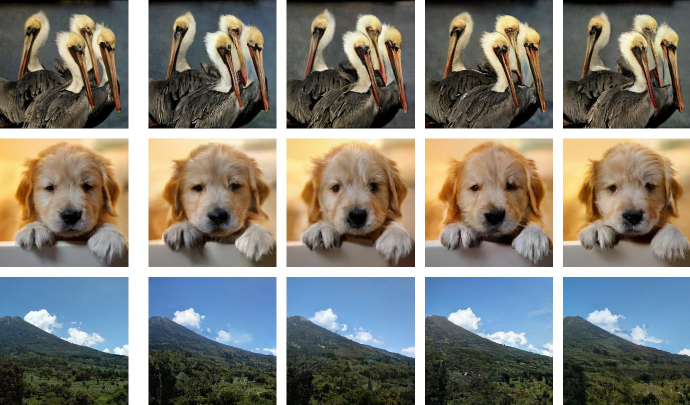}
    \caption{\label{fig:vaesamples}
    Samples from the VAE encoder distribution: The first column is an image from the ImageNet validation set. We encode it with the VAE encoder, obtain the approximate posterior distribution, and sample from it 4 times. The resulting 4 latent sequences are then decoded to images (last 4 columns). The semantic layout of the input image is well preserved, only the low-level textures change.
    }
\end{figure*}

\section{Additional visual examples} \label{sec:app:samples}

\begin{figure*}[hb]
    \centering
    \begin{tabularx}{\linewidth}{@{}XXX@{\hskip 1em}}
    GIVT-Causal \ours & GIVT-MaskGIT \ours & Validation Set
    \end{tabularx}
    \includegraphics[width=0.33\linewidth]{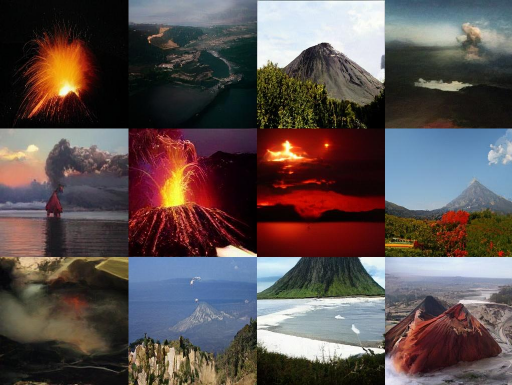}\hfill
    \includegraphics[width=0.33\linewidth]{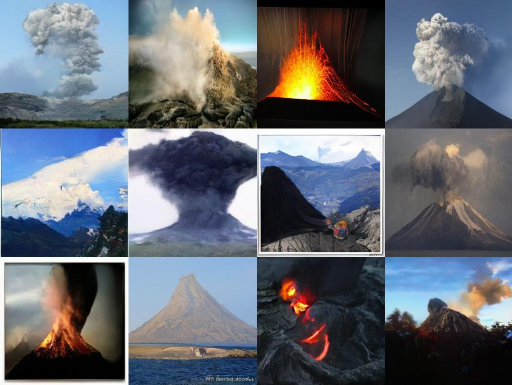}\hfill
    \includegraphics[width=0.33\linewidth]{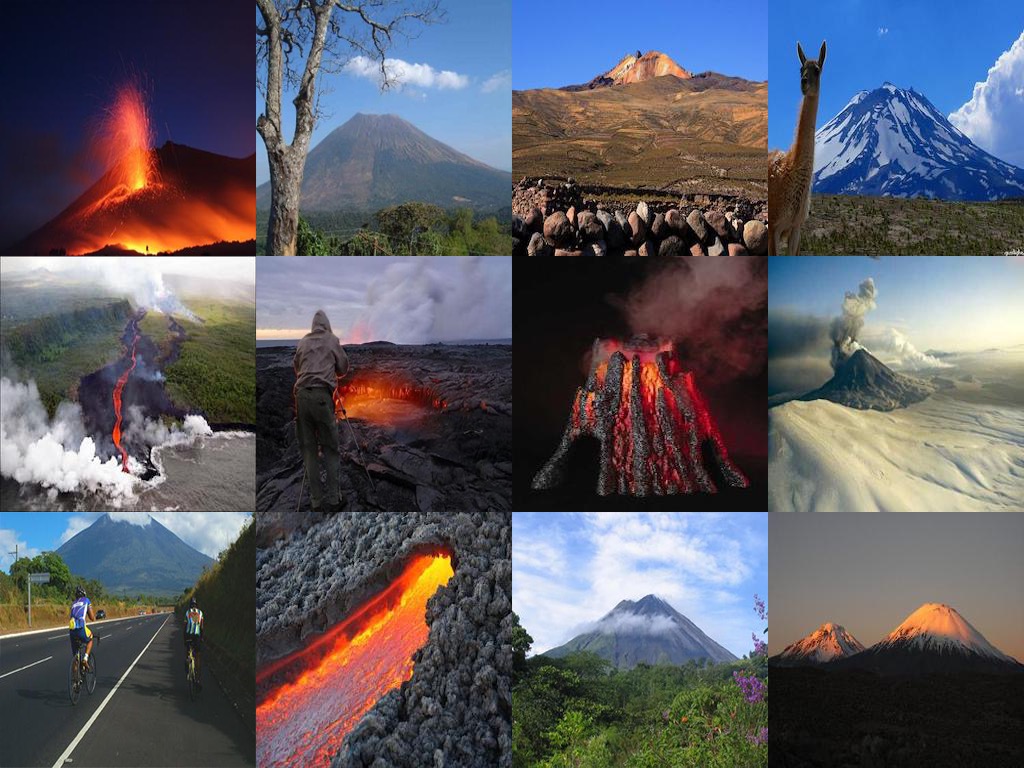} \\
    \begin{tabularx}{\linewidth}{@{}XXX@{\hskip 1em}}
    BigGAN~\cite{brock2018large} (via~\cite[Fig. 9]{chang2022maskgit}) & MaskGIT (VQ)~\cite[Fig. 9]{chang2022maskgit} & \hphantom{x}
    \end{tabularx}
    \includegraphics[width=0.33\linewidth]{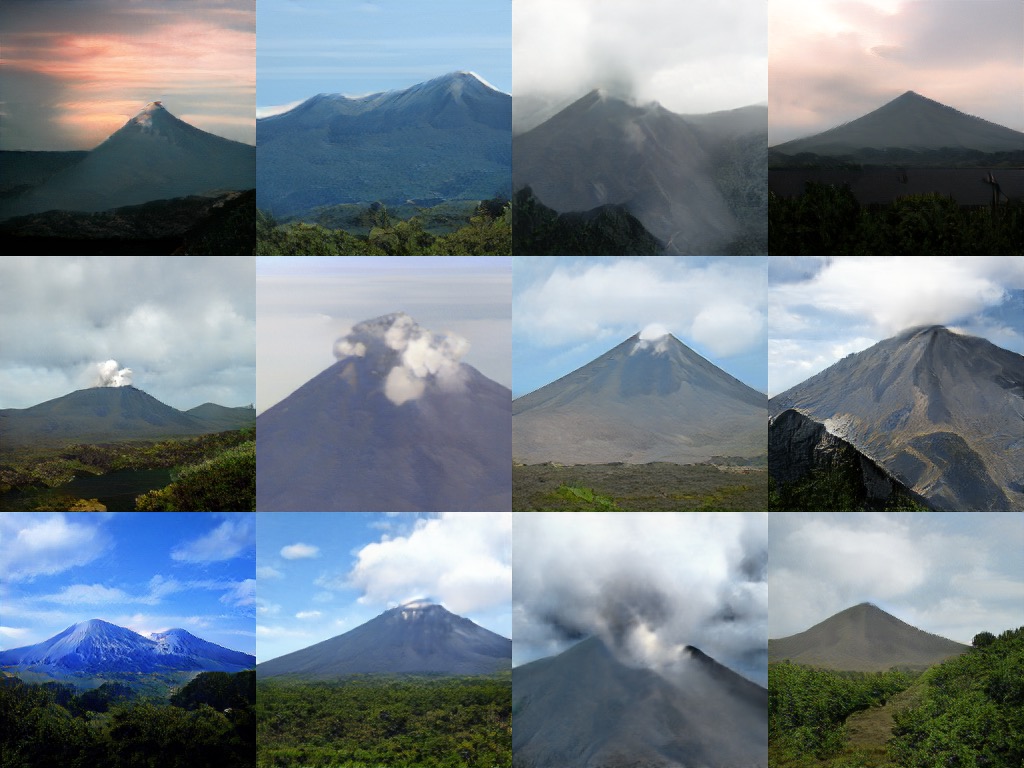}
    \includegraphics[width=0.33\linewidth]{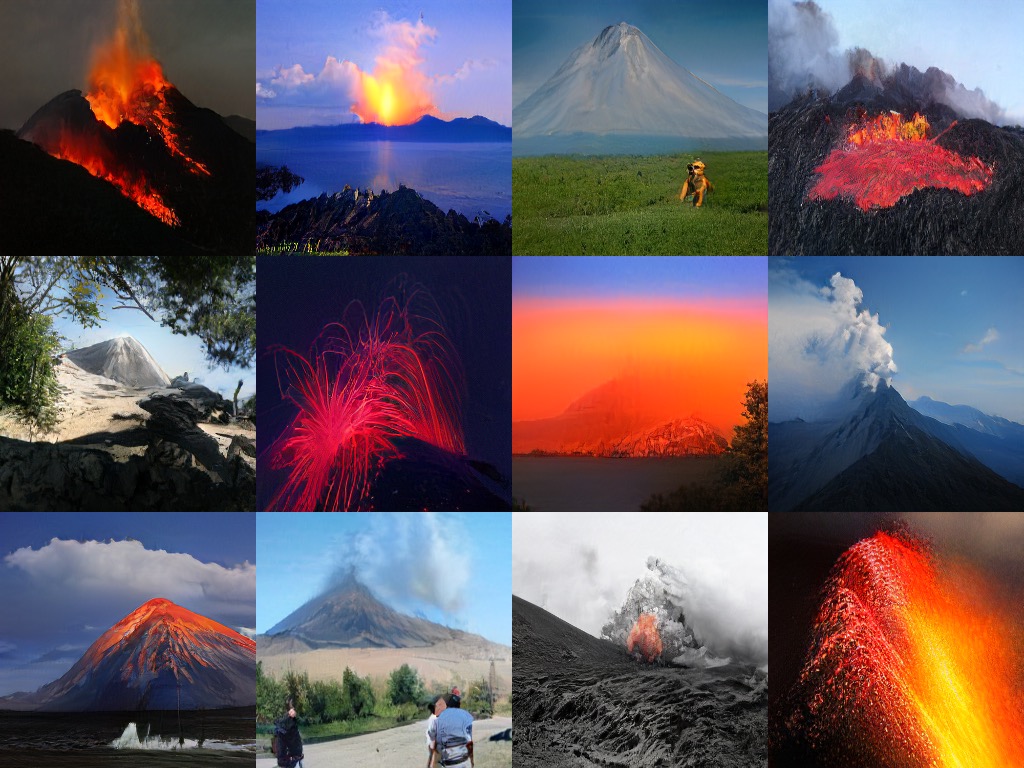} \phantom{\includegraphics[width=0.32\linewidth]{figs/diversity/maskgit_fig5.jpg}}
    \caption{\label{fig:diversity}Various samples for the \texttt{980} class to show the diversity of our samples (without DB-CFG). For reference, we copy examples from from~\cite[Fig.~9]{chang2022maskgit}.
    }
\end{figure*}

\begin{figure*}[ht]
    \centering
    \includegraphics[width=\linewidth]{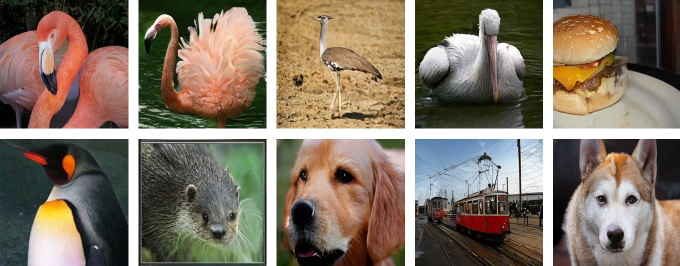}
    \caption{\label{fig:ar_sampling_256}Selected $256\times256$ samples from the GIVT-Causal ($t=0.95$, DB-CFG $=0.4$), for the same 10 ImageNet classes as in Fig.~\ref{fig:ar_sampling}.
    }
\end{figure*}

In Fig.~\ref{fig:vaesamples}, we show reconstructions from our VAE when feeding it images from the ImageNet validation set. We see low-level texture variations in the reconstructions when sampling from the encoder distribution multiple times.
Fig.~\ref{fig:diversity} shows multiple samples for a fixed label to show the diversity of our model and compare with baselines.
In Fig.~\ref{fig:ar_sampling_256} and \ref{fig:ar_sampling_256_l} we present samples from different $256 \times 256$ GIVT-Causal variants.
In Fig.~\ref{fig:masked_sampling} we show samples from MaskGIT for the same classes that we shown in Fig.~\ref{fig:ar_sampling}.
Fig.~\ref{fig:masked_sampling_512} shows samples from our $512\times512$ GIVT-MaskGIT model.
We explore changing labels midway through sampling (\ie, after generating the top half of the image) for GIVT-Causal in Fig.~\ref{fig:stitching}.

For our GIVT-Casual UViM models and VQ baselines, we show visual outputs in Figs.~\ref{fig:uvim:panoptic} and \ref{fig:uvim:depth}.

\begin{figure*}[ht]
    \centering
    \begin{tabularx}{\linewidth}{@{}X@{\hskip 2em}XXXX@{\hskip .5em}}
    VAE Sample & Step 1 & Step 4 & Step 8 & Step 16 \\
    \end{tabularx}
    \includegraphics[width=\linewidth]{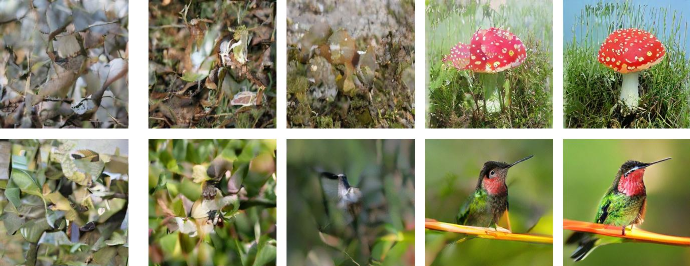}\vspace{-1ex}
    \caption{\label{fig:maskgit_sampling}
    \emph{First Column}: 
    Two images obtained when sampling from the VAE prior, resulting in a soup of low-level image textures. 
    \emph{Remaining Columns}:
    Visualizing the output of GIVT-MaskGIT, for two ImageNet classes (947, 94), after 1, 4, 8, 16 inference steps.
    As expected, the samples start to become more coherent as we perform more inference steps.
    \vspace{-1ex}
    }
\end{figure*}

\begin{figure*}[ht]
    \centering
    \includegraphics[width=\linewidth]{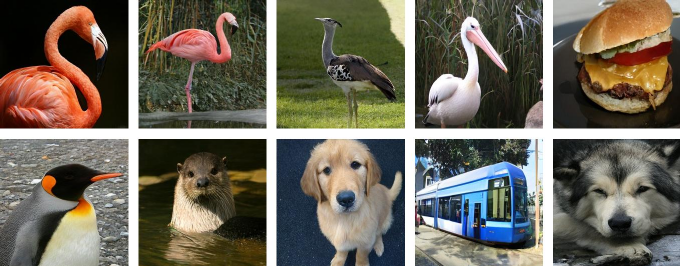}
    \caption{\label{fig:ar_sampling_256_l}Selected $256\times256$ samples from GIVT-Causal-L+A ($t=0.95$, DB-CFG $=0.4$), for the same 10 ImageNet classes as in Fig.~\ref{fig:ar_sampling}. 
    }
\end{figure*}

\begin{figure*}[ht]
    \centering
    \includegraphics[width=\linewidth]{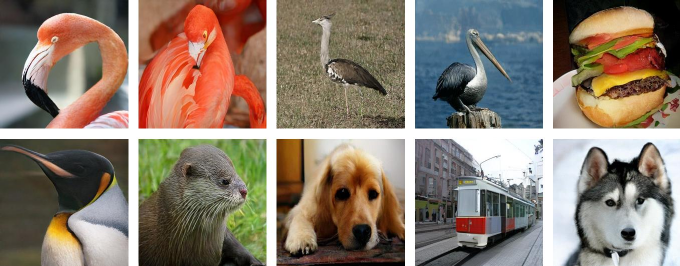}
    \caption{\label{fig:masked_sampling}Selected $256\times256$ samples from GIVT-MaskGIT ($t_C=60$, DB-CFG $=0.1$), for the same 10 ImageNet classes as in Fig.~\ref{fig:ar_sampling}. 
    }
\end{figure*}

\begin{figure*}[ht]
    \centering
    \includegraphics[width=\linewidth]{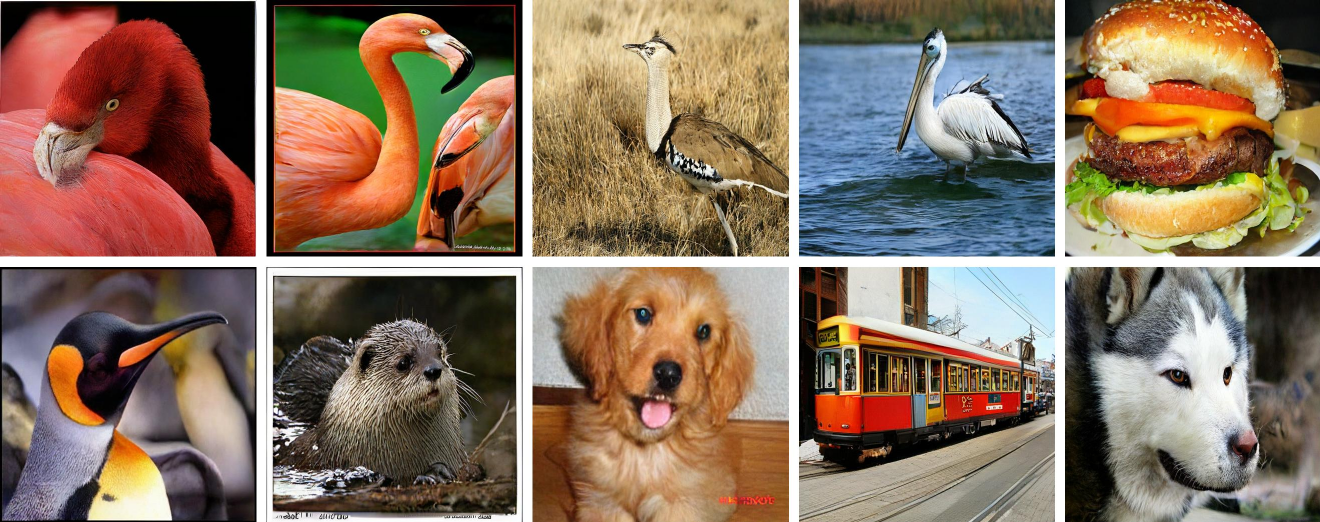}
    \caption{\label{fig:masked_sampling_512}Selected $512\times512$ samples from the GIVT-MaskGIT ($t_C=140$), for the same 10 ImageNet classes as in Fig.~\ref{fig:ar_sampling}. 
    }
\end{figure*}

\begin{figure*}[h]
    \centering
    \includegraphics[width=\linewidth]{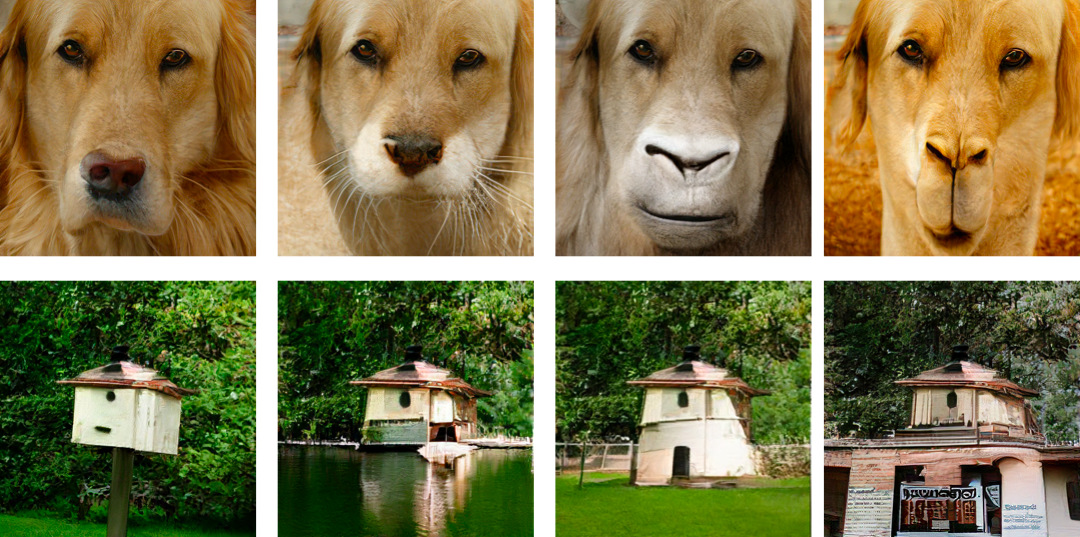}
    \caption{\label{fig:stitching}Changing the label half way through sampling (\ie, after the top half is generated) for GIVT-Causal: The first column uses the same label for the top and bottom half, the other columns switch to a different labels.
    Top row labels: ``golden retriever'' (207) for the top half; ``otter'' (360), ``gorilla'' (366), ``camel'' (355) for the bottom half.
    Bottom row labels: ``bird house'' (448) for the top half; ``boat house'' (449), ``light house'' (437), ``bakery'' (415) for the bottom half.
    Note that for each row, the top row \emph{latent} is always the same, but the overall color balance in the \emph{RGB output} might be different because of the VAE decoder (possibly due to GroupNorm layers).
    }
\end{figure*}

\begin{figure*}[h]
\vspace{-0.2cm}
    \begin{tabularx}{\linewidth}{@{}XXXX@{\hskip 0em}}
    Input & Ground Truth & GIVT-Causal UViM & VQ-based UViM
    \end{tabularx}
    \includegraphics[width=\linewidth]{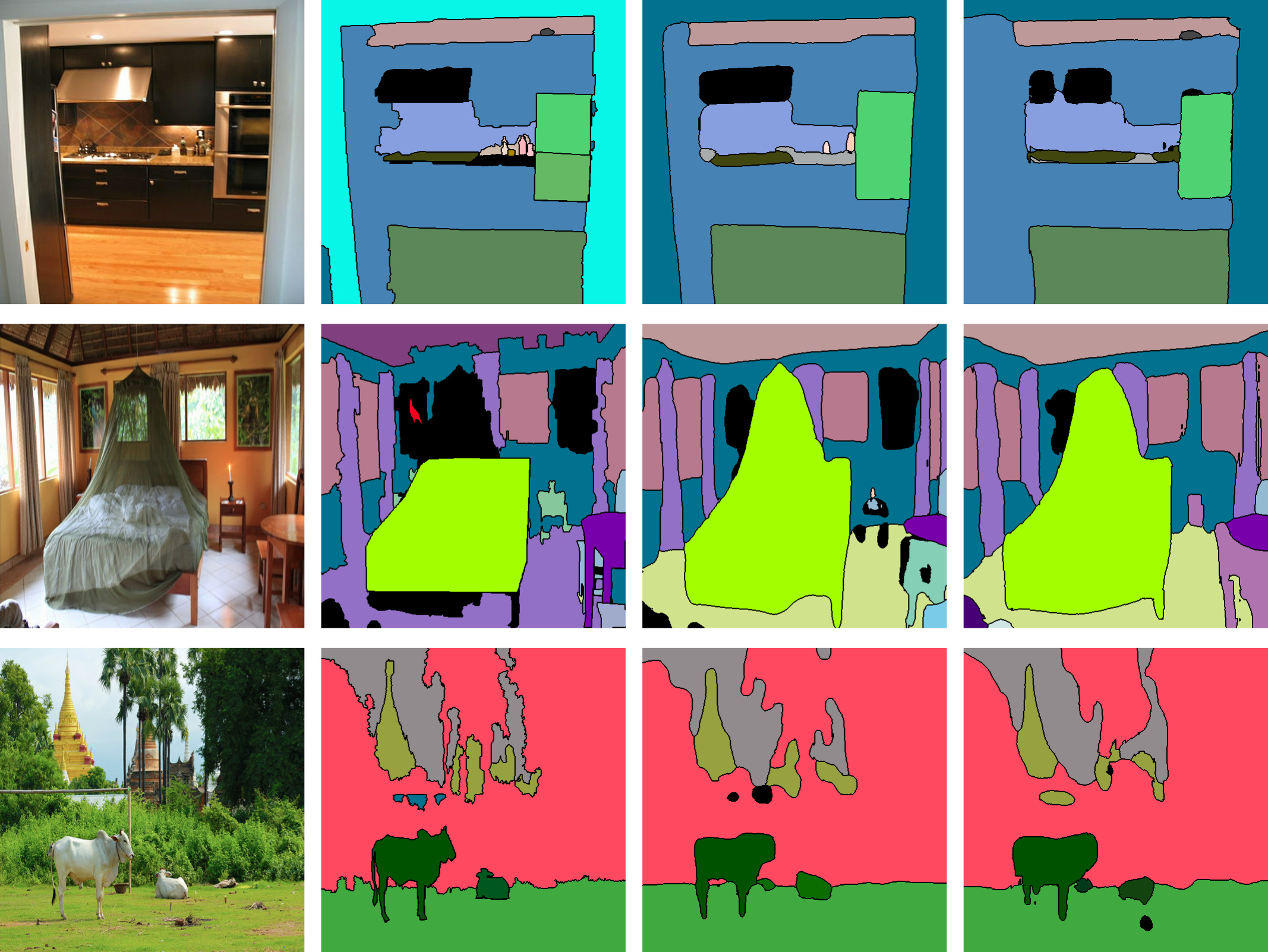}
    \caption{\label{fig:uvim:panoptic}
    `COCO Panoptic Segmentation' visual examples from our UViM models.
    Panoptic segmentation visual examples from COCO Panoptic 2017 for UViM based on GIVT-Causal and standard VQ-based UViM.
    \\
    }
    \centering
    \begin{tabularx}{\linewidth}{@{}XXXX@{\hskip 0em}}
    Input & Ground Truth & GIVT-Causal UViM & VQ-based UViM
    \end{tabularx}
    \includegraphics[width=\linewidth]{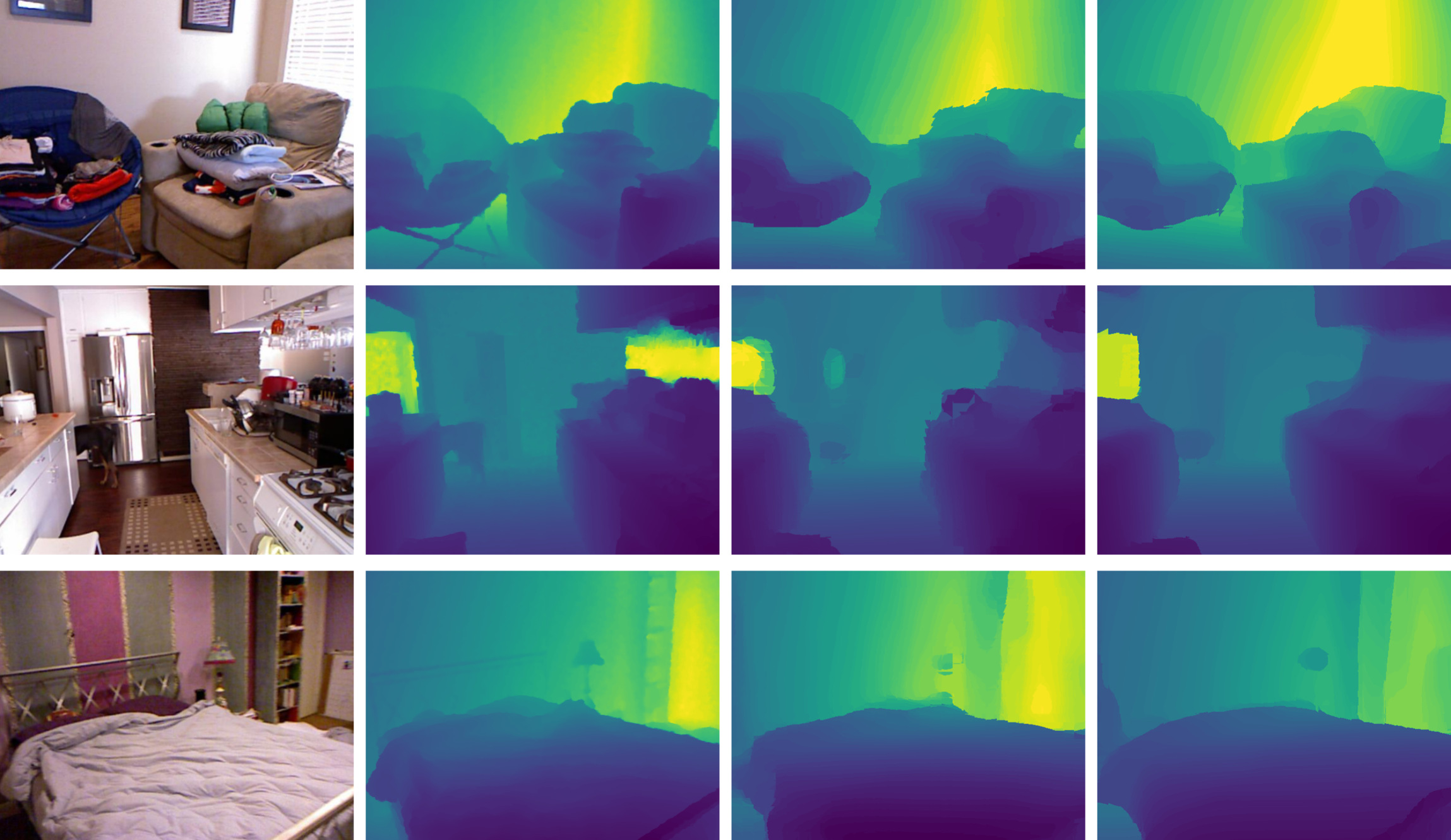}
    \caption{\label{fig:uvim:depth}
    Depth estimation visual examples from NYU Depth v2 for UViM based on GIVT-Causal and standard VQ-based UViM.
    }
\end{figure*}
\end{document}